\documentclass[10pt,journal,compsoc,utf8]{IEEEtran}
\usepackage{amsmath,amsfonts}

\usepackage[utf8]{inputenc}
\usepackage[numbers,sort&compress]{natbib}
\usepackage{graphicx}%
\usepackage{multirow}%
\usepackage{amsmath,amssymb,amsfonts}%
\usepackage{booktabs}%
\usepackage{url}
\usepackage{caption}
\usepackage[colorlinks,linkcolor=black,anchorcolor=black,citecolor=black]{hyperref}
\usepackage{makecell}
\usepackage{balance}
\usepackage{ragged2e}
\usepackage{colortbl}
\usepackage{tcolorbox}
\usepackage{comment}
\usepackage{adjustbox}
\usepackage{CJKutf8}

\usepackage{latexsym}
\usepackage{multicol}
\usepackage{booktabs}
\usepackage{hyperref}
\usepackage{graphicx}
\usepackage{subfigure}
\usepackage{fixltx2e}
\usepackage{algorithm}
\usepackage{algorithmic}
\usepackage{setspace}
\usepackage{soul}
\usepackage{amsmath}
\usepackage{amsthm}
\usepackage{amssymb}
\usepackage{pifont}
\newcommand{\cmark}{\ding{51}}%
\newcommand{\xmark}{\ding{55}}%
\usepackage[edges]{forest}

\ifCLASSINFOpdf
\else
\fi

\begin{document}
\begin{CJK}{UTF8}{gbsn}
%
\title{A Survey of Conversational Search}

%
%
%
%

\author{
\IEEEauthorblockN{Fengran Mo$^{1*}$, Kelong Mao$^{2*}$, Ziliang Zhao$^{2}$, Hongjin Qian$^{2}$, Haonan Chen$^{2}$, Yiruo Cheng$^{2}$, Xiaoxi Li$^{2}$ \\Yutao Zhu$^{2}$, Zhicheng Dou$^{2}$, Jian-Yun Nie$^{1}$}\\
    \IEEEauthorblockA{$^1$University of Montreal, Canada}
    \quad\IEEEauthorblockA{$^{2}$Renmin University of China}
\thanks{
\IEEEauthorblockA{\IEEEauthorrefmark{1}Equal Contribution.
}\\\IEEEauthorblockA{\IEEEauthorrefmark{2}We provide the comprehensive literature reading list in the following Github repository: \url{https://github.com/fengranMark/ConvSearch-Survey}.
}}
}
\IEEEtitleabstractindextext{%
\begin{justify}
\begin{abstract}
As a cornerstone of modern information access, search engines have become indispensable in everyday life. With the rapid advancements in AI and natural language processing (NLP) technologies, particularly large language models (LLMs), search engines have evolved to support more intuitive and intelligent interactions between users and systems. Conversational search, an emerging paradigm for next-generation search engines, leverages natural language dialogue to facilitate complex and precise information retrieval, thus attracting significant attention. Unlike traditional keyword-based search engines, conversational search systems enhance user experience by supporting intricate queries, maintaining context over multi-turn interactions, and providing robust information integration and processing capabilities. Key components such as query reformulation, search clarification, conversational retrieval, and response generation work in unison to enable these sophisticated interactions. In this survey, we explore the recent advancements and potential future directions in conversational search, examining the critical modules that constitute a conversational search system. We highlight the integration of LLMs in enhancing these systems and discuss the challenges and opportunities that lie ahead in this dynamic field. Additionally, we provide insights into real-world applications and robust evaluations of current conversational search systems, aiming to guide future research and development in conversational search.

\end{abstract}
\end{justify}

\begin{IEEEkeywords}
Conversational Search, Query Reformulation, Search Clarification, Conversational Retrieval and Generation, Domain-specific and User-centric Application, Benchmark and Evaluation
\end{IEEEkeywords}}

\maketitle

\IEEEdisplaynontitleabstractindextext

%
\IEEEpeerreviewmaketitle

\section{Introduction}\label{sec:introduction}

Search engines have become integral to modern society, serving as essential tools for fulfilling users' information needs. Their development has been significantly accelerated by advances in artificial intelligence (AI)~\cite{llmir_survey,mo2025conversational}.
Recently, with the rapid advancements in natural language processing (NLP) technologies, particularly the emergence of large language models (LLMs)~\cite{llm_survey}, search engines have evolved to offer more intelligent and interactive user experiences.
A notable advancement in this domain is conversational search, an emerging paradigm that provides natural language interactions for complex and accurate information access.
Commercial conversational AI search engines, such as Perplexity.ai and SearchGPT,\footnote{Perplexity.ai: \url{https://www.perplexity.ai/}, SearchGPT: \url{https://searchgpt.com/}}, have already been deployed, rapidly attracting a large user base and keeping significant growth. 
Conceptually, conversational search represents a paradigm shift towards interactive information access, which moves beyond single-shot queries to enable users to fulfill complex, often evolving, information needs through multi-turn natural language dialogue with a system. The core functionality of conversational search is to support information goal evolution, where a user's initial, possibly vague, need is refined or shifted through interaction. This necessitates a shared contextual understanding built turn-by-turn and often involves mixed initiative, where both the user and the system can guide the dialogue, e.g., the user asks or the system seeks clarification and proactively suggests refinements.
Ultimately, the search goal extends beyond mere relevance to encompass a richer user experience, focusing on efficiency, reducing cognitive load, enabling exploration, ensuring user satisfaction, and facilitating task completion~\cite{book23_CIS_survey}.

\begin{figure*}[tp]
\centering
\includegraphics[width=0.9\linewidth]{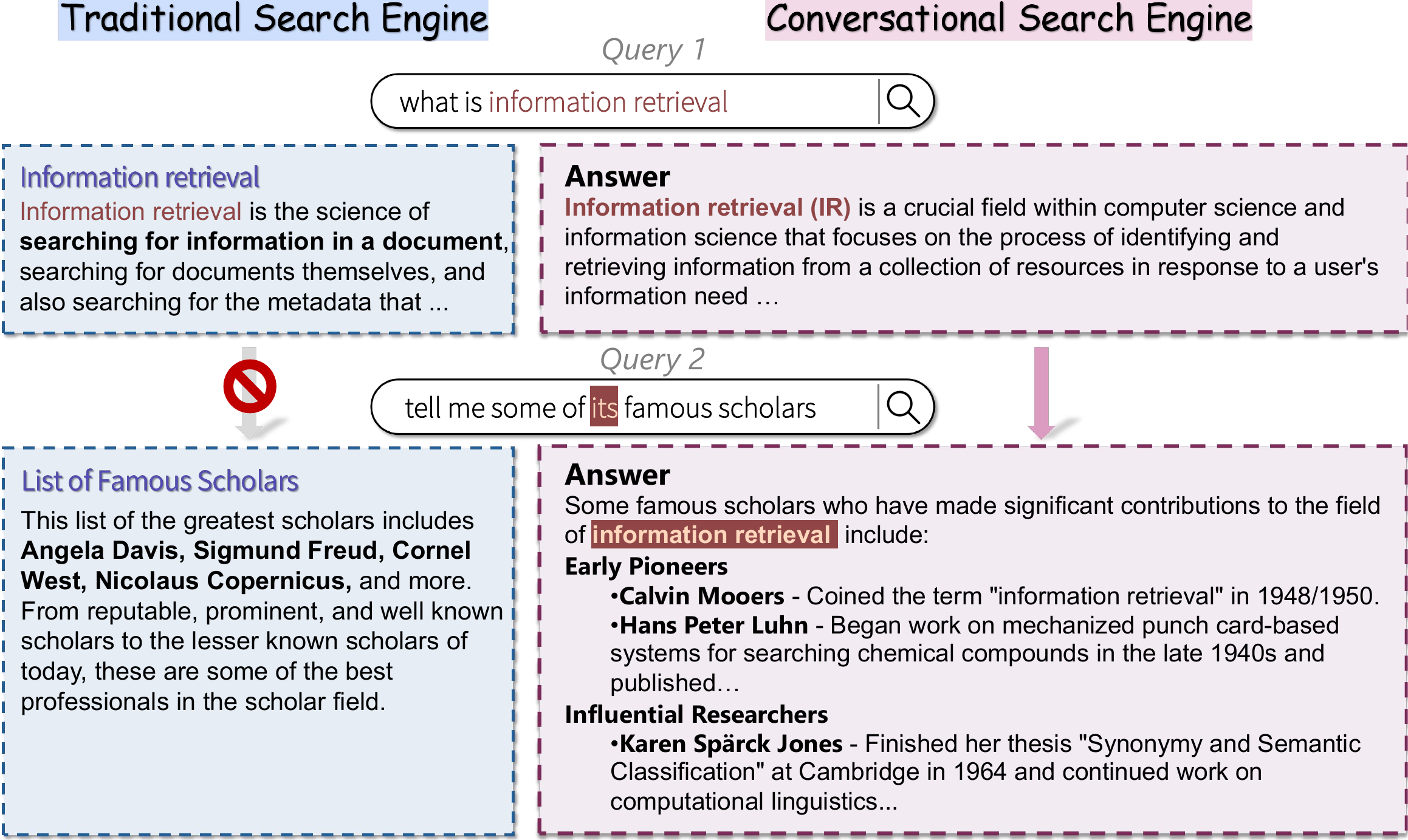}
\caption{Comparison between a traditional search engine and a new conversational search engine on an original and its follow-up query. The term ``its'' in the second query refers to the term ``information retrieval'' in the first query. However, it is challenging for traditional search engines to understand such omissions in follow-up user queries.}

\label{fig: google_vs_perplexity}
\end{figure*}

Compared to traditional search engines that rely on keywords or short phrases, conversational search leverages natural language dialogue for interactions~\cite{sigirforum18_cis_future, mao2022convtrans, mao2023large, mao2023search, mo2023convgqr}, greatly enhancing the efficiency of information exchange and optimizing the user experience. This approach supports more complex user queries, manages longer and more intricate historical contexts, and offers comprehensive capabilities for information integration and processing. Moreover, conversational search enables diverse methods of information delivery and possesses the ability to engage in active interactions with users~\cite{ren2021wizard, tacl23_inscit, aliannejadi2024trec}. For example, these systems can proactively ask clarifying questions~\cite{pyatkin2023clarifydelphi, mu2023clarifygpt, rahmani2024clarifying}, make recommendations~\cite{ai_open21_survey_of_conversational_recommendation}, and guide users in better expressing their needs through dialogue.
Fig.~\ref{fig: google_vs_perplexity} illustrates the differences between traditional search and conversational search.
In this real example, a traditional search engine (single-shot search) fails to maintain context between queries. After the user queries, ``What is information retrieval'', and then follows up with ``Tell me some of its famous scholars'', the search engine provides irrelevant results because it does not recognize that the user is still referring to ``information retrieval scholars''. In contrast, the conversational search engine (multi-turn search) successfully manages multiple query turns by retaining context, correctly interpreting the follow-up question, and providing accurate information about notable scholars in the field of information retrieval.

\begin{figure}[tp]
\includegraphics[width=.99\linewidth]{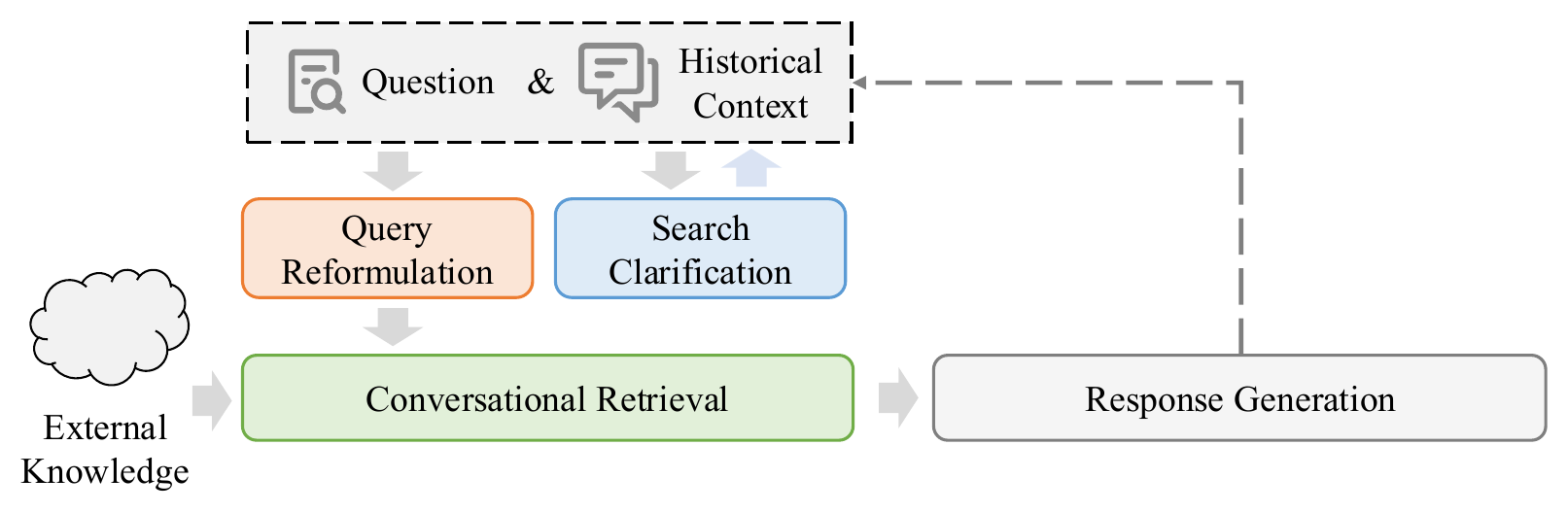}
\caption{A typical workflow for a conversational search system, which includes four key components: query reformulation, search clarification, conversational retrieval, and response generation.}
\label{fig: architecture}
\end{figure}

To conduct a systematic review of the technical advancements enabling this interactive paradigm, we examine four key functional capabilities typically identified in the literature as crucial for conversational search systems, including query reformulation, search clarification, conversational retrieval, and response generation, as illustrated in Fig.~\ref{fig: architecture}. These components work collaboratively to manage the multi-turn interactions, understand evolving intent, access information, and present it effectively, ultimately enhancing the overall search experience. 
It is important to note that recent advancements of powerful LLMs are progressively obscuring the boundaries between these traditionally distinct functions as the LLMs can often perform aspects of reformulation, retrieval, and generation in a more integrated, end-to-end fashion. 
This survey will explore both the developments within these established functional domains, given their continued relevance in contemporary systems and research, and analyze the implications of emerging integrative trends driven by LLMs with the attendant challenges and opportunities.


\noindent \textbf{Query Reformulation}. Query reformulation is a crucial initial step in both conversational search systems and traditional search engines. It involves various techniques such as query expansion~\cite{wang2023query2doc, emnlp23_query_rewriting_for_RAG}, rewriting~\cite{mo2023convgqr, yushi20}, and query decomposition~\cite{emnlp23_query_decomposition} to restructure the query, thereby enhancing the performance of subsequent system modules. In the context of conversational search, accurately interpreting the user's current information needs is particularly important, as it must be based on the ongoing conversation's evolving context. As conversations progress, this context can become increasingly complex and lengthy, posing challenges for traditional search engines, which often struggle with handling such multi-turn inputs. Therefore, query reformulation plays a vital role in distilling the entire conversational context, along with the current query, into a concise yet comprehensive representation of the user's current information needs. This process ensures that the subsequent components can more effectively process and respond to the user's query.

\noindent \textbf{Search Clarification}. Search clarification is another crucial component in conversational search systems. It enables users to refine their search queries through interactive dialogue, covering scenarios like seeking information, performing tasks, or navigating websites~\cite{aliannejadi2019asking, lee2023asking, feng2023towards}. Users often express their search intents ambiguously or in multifaceted ways. To improve understanding, the system can pose clarifying questions, such as ``Do you mean [specific term]?'' or ``Can you provide more details about [topic]?'', rather than directly answering the query. When the system detects that clarification is needed, it proactively asks these questions to understand the user's intent better. This approach ensures that the system provides more accurate and relevant search results, enhancing the overall user experience by making the search process more personalized and effective. The primary challenges in search clarification include accurately and efficiently identifying when clarification is necessary and generating appropriate clarifying questions.

\noindent \textbf{Conversational Retrieval}. After the query and context have been reformulated and any necessary clarifications have been made, the system proceeds to retrieve relevant information from external knowledge bases to satisfy the user's information needs. Unlike traditional ad-hoc retrieval, conversational retrieval faces the unique challenge of extracting useful knowledge chunks from vast knowledge bases within the context of an ongoing, complex dialogue. This process requires the system to effectively manage the intricacies of multi-turn interactions and extended context lengths to ensure that the most relevant information is retrieved, enabling accurate and helpful responses for the user~\cite{ORConvQA}. A straightforward approach is to leverage the previously reformulated conversational query for retrieval~\cite{mo2023convgqr,mao2023large}. However, this method may fail because upstream query rewriters often struggle to optimize based on downstream retrieval signals. Moreover, as the context grows longer and the user's information needs become more complex, generating a concise yet effective query rewrite becomes increasingly challenging. An emerging and promising solution to this problem is conversational dense retrieval, which involves training a session encoder to directly encode the entire context~\cite{jeong2023phrase,coted,ConvAug,ChatRetriever,HAConvDR}. By avoiding explicit query rewriting, this method can utilize ranking signals for direct optimization, potentially leading to superior performance. Nonetheless, training an efficient conversational dense retriever remains challenging due to the complexity of encoding the entire context. Additionally, similar to traditional search engines, conversational search systems need to re-rank the retrieved content more precisely within the complex context~\cite{ConvRerank}. This re-ranking process is crucial, as it ensures that the subsequent response generation module has access to the most relevant content, thereby enabling the generation of more accurate and helpful responses.

\noindent \textbf{Response Generation}. A key distinction between next-generation conversational search systems and traditional search engines is their ability to deliver more than just a list of links. Conversational search systems can tailor their responses to better suit users' needs, providing various output formats such as direct and concise answers, summarized contents~\cite{arxiv24_memorag}, or even structured data like tables~\cite{acl22_mmcoqa}. After the conversational retrieval and re-ranking stages, the system tries to synthesize the retrieved information along with the contextual dialogue to generate precise and relevant responses. However, this generation process still faces a lot of challenges. These include determining the appropriate format for presenting the content, optimizing model fidelity to effectively utilize retrieved knowledge, managing conflicts between internal and external knowledge sources, handling extremely long contextual information during generation~\cite{arxiv24_ultragist}, and providing accurate citation labeling to facilitate source verification~\cite{emnlp23_ALCE}. Addressing these challenges is also important for enhancing the effectiveness and reliability of conversational search systems. 

The four key components—query reformulation, search clarification, conversational retrieval, and response generation—constitute the foundational framework of a general conversational search system. Beyond these core elements, such systems have been successfully adapted to specialized domains, including healthcare~\cite{xia2022medconqa, bhowmik2021fast, han2023medalpaca, labrak2024biomistral, toma2023clinical, bao2023disc, pal2023med, wang2024jmlr, jin2024better, alonso2024medexpqa}, finance~\cite{sharma2021stockbabble, deng2022pacific, liu2023tab, chen2022convfinqa, choi2023conversational, xie2024pixiu}, and legal fields~\cite{liu2021conversational, slocum2016conversational, liu2023investigating, askari2023closer}, where they are tailored to meet specific user needs and contextual requirements. These applications encompass a wide range of fields, from personalized medical consultations that utilize advanced language models~\cite{xia2022medconqa} to intuitive financial advisory services that assist investors in navigating complex market data~\cite{chen2022convfinqa}. In the legal domain, conversational search systems facilitate efficient case retrieval and support nuanced legal reasoning~\cite{liu2021conversational}. Additionally, beyond these specialized areas, conversational search systems are enhancing e-commerce platforms by enabling personalized product recommendations and responsive customer service interactions~\cite{jia2022convrec}. These varied applications highlight the adaptability and increasing significance of conversational search in addressing contemporary information retrieval challenges.

\definecolor{fill_0}{RGB}{251, 245, 251}
\definecolor{fill_1}{RGB}{246, 249, 255}
\definecolor{fill_2}{RGB}{244, 249, 241}

\definecolor{draw_0}{RGB}{168, 80, 168}
\definecolor{draw_1}{RGB}{54, 110, 210}
\definecolor{draw_2}{RGB}{112, 173, 71}

\definecolor{fill_leaf}{RGB}{248, 248, 248}
\definecolor{draw-leaf}{RGB}{135, 135, 135}

\tikzstyle{my-box}=[
    rectangle,  
    draw=draw-leaf,
    rounded corners,
    text opacity=1,
    minimum height=1.5em,
    minimum width=5em,
    inner sep=2pt,
    align=center,
    fill opacity=.5,
    line width=0.8pt,
]
\tikzset{
leaf/.style={
my-box,
minimum height=1.5em,
fill=fill_leaf, 
text=black,
align=left,
font=\footnotesize,
inner xsep=2pt,
inner ysep=4pt,
line width=1pt
}
}
\begin{figure*}[!t]
\centering
\begin{adjustbox}{width=0.73\textwidth}
\begin{forest}
forked edges,
for tree={
    grow=east,
    reversed=true,
    anchor=base west,
    parent anchor=east,
    child anchor=west,
    base=center,
    font=\large,
    rectangle,
    draw=draw-leaf,
    rounded corners,
    align=left,
    text centered,
    minimum width=5em,
    edge+={darkgray, line width=1pt},
    s sep=3pt,
    inner xsep=2pt,
    inner ysep=3pt,
    line width=1.2pt,
    ver/.style={rotate=90, child anchor=north, parent anchor=south, anchor=center},
},
where level=0{text width=10.5em,fill=fill_0,draw=draw_0,font=\large,}{},
where level=1{text width=10.5em,fill=fill_1,draw=draw_1,font=\normalsize,}{},
where level=2{text width=14em,fill=fill_2,draw=draw_2,font=\normalsize,}{},
where level=3{font=\footnotesize,}{},
[
    Conversational \\ Search, minimum height=2.8em
    [ 
        Query Reformulation \\(Sec~\ref{sec: Query Reformulation}) 
        [
            Conversational Query \\ Reformulation Models (Sec~\ref{sec: CQR model})
            [
                \textbf{\textit{Query Expansion}}: MISE~\cite{kumar2020making}{,} QuReTeC~\cite{voskarides2020query}{,} CQE~\cite{lin2021contextualized}{,} TPCS~\cite{mele2020topic}{,} \\ ConvRelExpand~\cite{ConvRelExpand}, leaf, text width=27.8em
            ]
            [
                \textbf{\textit{Query Rewriting}}: ConvQR~\cite{yu2020few}{,} T5QR~\cite{lin2020conversational}{,} Transformer++~\cite{transformerplus}{,} \\ QRODCQA~\cite{del2021question}{,} QRCPR~\cite{vakulenko2021comparison}{,} CONQRR~\cite{wu2022conqrr}{,} RLCQR~\cite{chen2022RLCQR}{,} \\ ExpCQR~\cite{hongjin22}{,} EDIRCS~\cite{mao2023search}{,} LLM4CS~\cite{mao2023large}{,} LLM-Aided~\cite{ye2023enhancing}{,} IterCQR~\cite{jang2023itercqr}, leaf, text width=27.8em
            ]
            [
                \textbf{\textit{Hybrid}}: ConvGQR~\cite{mo2023convgqr}{,} LLM4CS~\cite{mao2023large}{,} RETPO~\cite{yoon2024ask}{,} CHIQ~\cite{mo2024chiq}, leaf, text width=27.8em
            ]
        ]
        [
            Dataset (Sec~\ref{sec: QR Analysis}) and \\ Evaluation (Sec~\ref{sec: QR assess})
            [
                \textbf{\textit{Dataset Analysis}}: CANARD~\cite{elgohary2019can}{,} QReCC~\cite{anantha2021open}{,} TREC-CAsT~\cite{dalton2020cast,dalton2021cast,dalton2022cast,owoicho2022trec}, leaf, text width=27.8em
            ]
            [
                \textbf{\textit{Evaluation}}: Lexical Overlapping~\cite{elgohary2019can,anantha2021open}{,} End-to-End~\cite{ConvRelExpand,mao2023search}, leaf, text width=27.8em
            ]
        ]
    ]
    [ 
        Search Clarification\\ (Sec \ref{sec:clari}) 
        [
            Clarification for Conversational \\ Retrieval (Sec \ref{sec:clari-cs}) 
            [
                \textbf{\textit{Clarifying Question Selection}}: NeuQS~\cite{aliannejadi2019asking}{,} GT~\cite{hashemi2020guided}{,} MRR-BERT~\cite{bi2021asking}{,} \\ BERT-C-cq \& BERT-C-P-cq~\cite{mass2022conversational}, leaf, text width=27.8em
            ]
            [
                \textbf{\textit{Clarifying Question Generation}}: QF-GPT~\cite{sekulic2021towards}{,} ZSFC~\cite{wang2023zero}, leaf, text width=27.8em
            ]
            [
                \textbf{\textit{Extension Studies}}: User Simulation~\cite{wang2024depth}{,} Multi-modal~\cite{yuan2024asking}, leaf, text width=27.8em
            ]
        ]
        [
            Web Search Clarification \\ (Sec \ref{sec:clari-web}) 
            [
                \textbf{\textit{Generating Clarifying Questions}}: RTC \& QLM \& QCM~\cite{zamani2020generating}{,} RLC~\cite{zamani2020analyzing}{,} \\ TG-ClariQ~\cite{wang2021template}{,} SRQG~\cite{zhao2022generating}{,} BART-structure~\cite{zhao2023improving}, leaf, text width=27.8em
            ]
            [
                \textbf{\textit{Generating Candidate Facets}}: QDMiner~\cite{dou2011finding}{,} QFI \& QFJ~\cite{kong2013extracting, kong2014extending}{,} \\ NMIR~\cite{hashemi2021learning}{,} PINMIR~\cite{hashemi2022stochastic}{,} PLM-based~\cite{samarinas2022revisiting}{,} BART-structure~\cite{zhao2023improving}{,} \\ Coherency~\cite{litvinov2024analyzing}{,} Multi-turn~\cite{zhao2024generating}{,} Seq~\cite{ni2023comparative}{,} LLM Editing~\cite{lee2024enhanced}, leaf, text width=27.8em
            ]
            [
                \textbf{\textit{Extension Studies}}: RPR \& EQG \& LLMG~\cite{liu2024mining}{,} ELBERT~\cite{sekulic2021user}{,} \\ CQ-Rank~\cite{lotze2021ranking}{,} GACS~\cite{gao2022search}, leaf, text width=27.8em
            ]
        ]
        [
            Search Clarification for \\ Question Answering (Sec \ref{sec:clari-qa}) 
            [
                \textbf{\textit{Community QA}}: EVPI~\cite{rao2018learning}{,} GAN-utility~\cite{rao2019answer}{,} SimQ~\cite{trienes2019identifying}{,} SBERT~\cite{kumar2020ranking}, leaf, text width=27.8em
            ]
            [
                \textbf{\textit{Knowledge-based QA}}: HAN \& DMN~\cite{xu2019asking}{,} VP \& PP~\cite{nakano2022pseudo}{,} \\ Tree-of-clarification~\cite{kim2023tree}{,} CAmbigNQ~\cite{lee2023asking}, leaf, text width=27.8em
            ]
        ]
        [
            Domain-specific Search \\ Clarification (Sec \ref{sec:clari-domain}) 
            [
                \textbf{\textit{Conversational Recommender System}}: Conversation Policy~\cite{christakopoulou2016towards}{,}\\ DeepCRS~\cite{li2018towards}{,} DialogueCRS~\cite{sun2018conversational}{,} Knowledge CRS~\cite{chen2019towards}{,}\\ Topic-guided~\cite{zhou2020towards}{,} KG-enhanced~\cite{zhou2020improving}{,} SCPR~\cite{lei2020interactive}{,} EAR~\cite{lei2020estimation}{,}\\ UNICORN~\cite{deng2021unified}{,} MCMIPL~\cite{zhang2022multiple}{,} RecLLM~\cite{friedman2023leveraging}{,} iEvalLM~\cite{wang2023rethinking}, leaf, text width=27.8em
            ]
            [
                \textbf{\textit{Legal}}: CvsT~\cite{liu2021conversational}{,} QGBM~\cite{liu2022query}{,} LeClari~\cite{liu2023leveraging}, leaf, text width=27.8em
            ]
            [
                \textbf{\textit{Other}}: interactive classification~\cite{zamani2020generating}{,} image guessing~\cite{white2021open}{,}\\ 
                product description  ~\cite{majumder2021ask, zhang2021diverse}{,} collaborative building~\cite{shi2022learning}{,}\\ programming~\cite{eberhart2022generating, li2023python}{,} user simulation~\cite{erbacher2022interactive}{,}\\ social or moral situation~\cite{pyatkin2023clarifydelphi}{,} task-oriented dialogues~\cite{feng2023towards}{,} math~\cite{mansouri2023clarifying}{,}\\ spatial  reasoning~\cite{deng2023learning}{,} collaborative dialogue~\cite{testoni2024asking}, leaf, text width=27.8em
            ]
        ]
        [
            Search Clarification \\ with LLMs  (Sec \ref{sec:clari-llm}) 
            [
                ClarifyGPT~\cite{mu2023clarifygpt}{,} GATE~\cite{li2023eliciting}{,} STaR-GATE~\cite{andukuri2024star}{,} RaR~\cite{deng2023rephrase}, leaf, text width=27.8em
            ]
        ]
    ]
    [
        Retrieval (Sec \ref{sec:retrieval})
        [
            Conversation \\ Modeling  (Sec \ref{subsec:Conversation_Modeling})
            [
                HAM~\cite{ham}{,} ORConvQA~\cite{ORConvQA}{,} CoSPLADE~\cite{CoSPLADE}{,} DHM~\cite{DHM}{,} \\ DGRCoQA~\cite{DGRCoQA}{,} Conv-CoA~\cite{Conv-CoA}{,} ChatRetriever~\cite{ChatRetriever}, leaf, text width=27.8em
            ]
        ]
        [
            Context Denoising\\ (Sec \ref{subsec:Context_Denoising})
            [
                \textbf{\textit{Implicit}}: ConvDR~\cite{ConvDR}{,} COTED~\cite{coted}{,} ZeCo~\cite{zeco}{,} QRACDR~\cite{QRACDR}{,}\\ DiSCO~\cite{lupart2025disco}, leaf, text width=27.8em
            ]
            [
                \textbf{\textit{Explicit}}:
                ConvRelExpand~\cite{ConvRelExpand}{,} HAConvDR~\cite{HAConvDR}, leaf, text width=27.8em
            ]
        ]
        [
            Data Augmentation\\ (Sec \ref{subsec:Data_Augmentation})
            [
                \textbf{\textit{Relevance Judgment}}: CQE~\cite{CQE}{,} InstructoR~\cite{InstructoR}, leaf, text width=27.8em
            ]
            [
                \textbf{\textit{Contrastive Sample}}:
                HAConvDR~\cite{HAConvDR}{,} Shortcut~\cite{Shortcut}, leaf, text width=27.8em
            ]
            [
                \textbf{\textit{Conversation Session}}: ConvTrans~\cite{ConvTrans}{,} ConvSDG~\cite{ConvSDG}{,} \\ CONVERSER~\cite{huang2023converser}{,} ConvAug~\cite{ConvAug}, leaf, text width=27.8em
            ]
        ]
        [
            Interpretation  (Sec \ref{subsec:Interpretation_CDR})
            [
             LeCoRE~\cite{Lecore}{,} Explaignn~\cite{christmann2023explainable}{,} CONVINV~\cite{ConvInv}, leaf, text width=27.8em
            ]
        ]
        [
            Re-ranking  (Sec \ref{subsec:Reranking_CS})
            [
                MonoQA~\cite{monoQA}{,} ConvRerank~\cite{ConvRerank}{,} CoSPLADE~\cite{CoSPLADE}{,} ConvSim~\cite{ConvSim}, leaf, text width=27.8em
            ]
        ]
    ]
    [
        Generation (Sec \ref{sec: generation})
        [
            Historical Search\\ Results (Sec~\ref{sec: history search})
            [
                 Fine-Grained RAG~\cite{ye2024boosting}{,} Conv-CoA~\cite{Conv-CoA}{,} EvoRAG~\cite{cheng2025evolving}, leaf, text width=27.8em
            ]
        ]
        [
            Context Dependency\\ Modeling (Sec~\ref{subsec:context_dependency_modeling}) 
            [
                ConvRelExpand~\cite{mo2023learning_to_relate}{,} LeCoRE~\cite{mao2023LeCoRE}{,} HAConvDR~\cite{mo2401HAConvDR}, leaf, text width=27.8em
            ]
        ]
        [
            Conversational Knowledge\\ Attribution (Sec~\ref{subsec:knowledge_attribution})
            [
                WebGPT~\cite{webgpt}{,} LaMDA~\cite{lamda}{,} WebBrain~\cite{webbrain}{,} UniConv~\cite{mo2025uniconv}{,} Coral~\cite{cheng2024coral} , leaf, text width=27.8em
            ]
        ]
    ]
    [
        Domain-specific and \\ User-centric (Sec \ref{sec:domain})
        [
            Domain-specific (Sec \ref{subsec:domain-specific})
            [
                \textbf{\textit{Medical Domain}}: 
                Clinical Camel~\cite{toma2023clinical}{,} 
                DISC-MedLLM~\cite{bao2023disc}{,} \\Zhongjing~\cite{yang2024zhongjing}{,} MedConQA~\cite{xia2022medconqa}{,} BioMistral~\cite{labrak2024biomistral}{,} Med-HALT~\cite{pal2023med} 
                , leaf, text width=27.8em
            ]
            [
                \textbf{\textit{Financial Domain}}: 
                StockBabble~\cite{sharma2021stockbabble}{,} PACIFIC~\cite{deng2022pacific}{,} Tab-CQA~\cite{liu2023tab}{,}\\ ConvFinQA~\cite{chen2022convfinqa}{,} ConFIRM~\cite{choi2023conversational}{,} FinMA~\cite{xie2024pixiu}, leaf, text width=27.8em
            ]
            [
                \textbf{\textit{Legal Domain}}: Ratiojuris~\cite{slocum2016conversational}{,} Search Behavior~\cite{liu2021conversational}{,} ConvLegal~\cite{liu2022query}{,} \\ ConvAA~\cite{liu2023investigating}{,} CLosER~\cite{askari2023closer},leaf, text width=27.8em
            ]
            [
                \textbf{\textit{Other Domains}}:
                E-ConvRec~\cite{jia2022convrec}{,} MG-ShopDial~\cite{bernard2023mg}{,} ConvENKG~\cite{zamiri2024benchmark}{,} \\MMConv~\cite{liao2021mmconv}{,} MMCoQA~\cite{li2022mmcoqa}{,} MoqaGPT~\cite{zhang2023moqagpt}, leaf, text width=27.8em
            ]
        ]
        [
            User-centric (Sec \ref{subsec:user-centric})
            [
                ConvBehavior~\cite{chu2022convsearch}{,} PAQA~\cite{erbacher2024paqa}{,} ProConvAgent~\cite{deng2024towards}{,} TITAN~\cite{yan2023titan}{,} \\LAPS~\cite{joko2024doing}{,} PersonalizedCIR~\cite{mo2024leverage}{,} ConvWebAgent~\cite{deng2024multi}{,} Conv2Q~\cite{meng2025bridging}{,}\\AdaptivePCIR~\cite{mo2025personalized} , leaf, text width=27.8em
            ]
        ]
    ]
    [
        Benchmark and \\Evaluation (Sec \ref{sec:evaluation})
        [
            Evaluation (Sec \ref{subsec:evaluation-overview})
            [
                \textbf{\textit{Retrieval-based}}:
                EvalCranfield~\cite{fu2022evaluating}{,} Ditch~\cite{li2022ditch}{,} RobustEval~\cite{siblini2021towards}{,}\\ RefineUS~\cite{salle2021studying}{,} CSOffline~\cite{lipani2021doing}{,} SRisk~\cite{wang2022simulating}{,} ResponseUS~\cite{wang2023depth}{,}\\ ConvSim~\cite{owoicho2023exploiting}{,} USEIAS~\cite{balog2023user}{,} USConvAgent~\cite{bernard2024leveraging}{,} USMix~\cite{sekulic2022evaluating}{,}\\ SIPCIS~\cite{meng2023system}{,} PACS~\cite{fu2023priming},leaf, text width=27.8em
            ]
            [
                \textbf{\textit{Generation-based}}:
                eRAG~\cite{salemi2024evaluating}{,} PRGR~\cite{lajewska2023towards}{,} EvalLLMQA~\cite{alinejad2024evaluating} ,leaf, text width=27.8em
            ]
            [
                \textbf{\textit{User-based}}:
                URS~\cite{wang2024user}{,} CCPCS~\cite{trippas2024re}{,} DMCBSDS~\cite{ji2024towards} ,leaf, text width=27.8em
            ]
        ]
        [
            Benchmark (Sec \ref{sec:evaluation})
            [
                MSDialog~\cite{qu2018analyzing}{,} TREC CAsT-19~\cite{dalton2020cast}{,} Qulac~\cite{aliannejadi2019asking}{,} CANARD~\cite{elgohary2019can}{,}\\
                MANtIS~\cite{penha_mantis}{,} WoW~\cite{dinan_iclr19_wizard}{,} CLAQUA~\cite{xu2019asking}{,} TREC CAsT-20~\cite{dalton2021cast}{,}\\
                OR-QuAC~\cite{qu2020open}{,} MIMICS~\cite{zamani2020mimics}{,} ClarQ~\cite{kumar2020clarq}{,} ClariQ~\cite{aliannejadi2020convai3}{,}\\
                ClariQ-FKw~\cite{aliannejadi2020convai3}{,} WISE~\cite{ren2021wizard}{,} QReCC~\cite{anantha2021open}{,} TopiOCQA~\cite{adlakha2022topiocqa}{,}\\
                TREC CAsT-21~\cite{dalton2022cast}{,} MultiDoc2Dial~\cite{feng_emnlp21_multidoc2dial}{,} Abg-CoQA~\cite{guo2021abg}{,} SaaC~\cite{ren_serp_saac_case}{,}\\
                ConvS~\cite{chu2022convsearch}{,} TREC CAsT-22~\cite{owoicho2022trec}{,} MIMICS-Duo~\cite{tavakoli2022mimics}{,} Doc2Bot~\cite{fu_emnlp22_doc2bot}{,}\\
                FOCUSQA~\cite{Barlacchi_emnlp22_focusqa}{,} TREC iKAT-23~\cite{aliannejadi2024trec}{,} ConvRAG~\cite{ye2024boosting}{,} Coral~\cite{cheng2024coral}\\
                StatCan Dialogue~\cite{lu_eacl23_StatCan_Dialogue_Dataset}{,} DomainRAG~\cite{wang2024richrag}{,} ProCIS~\cite{samarinas2024procis}{,} Melon~\cite{yuan2024asking}, leaf, text width=27.8em
            ]
        ]
    ]
]
\end{forest}
\end{adjustbox}
\caption{A structured taxonomy of conversational search systems which categorizes existing research.}
\vspace{-4ex}
\label{fig:taxonomy}
\end{figure*}
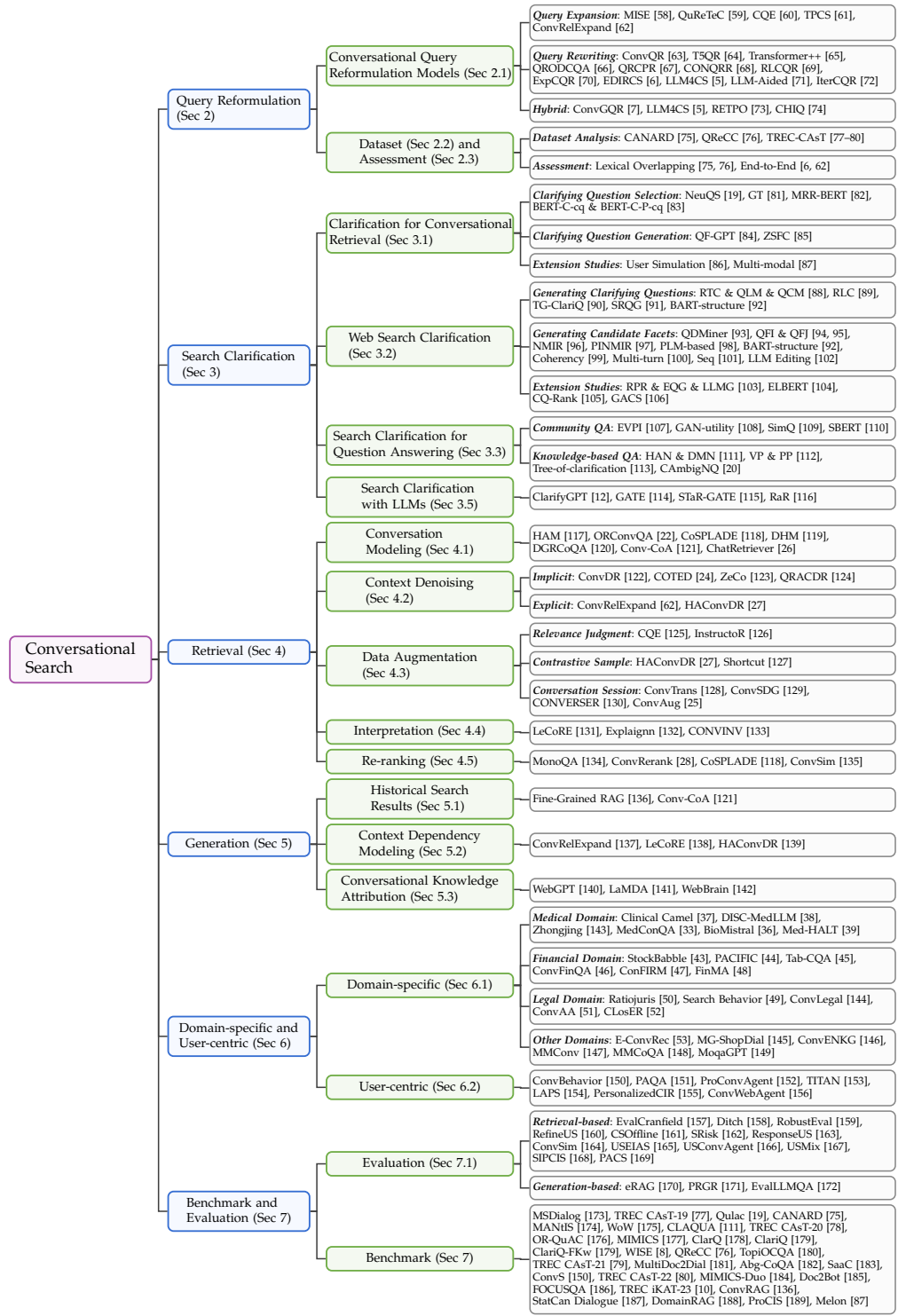

In conclusion, this survey aims to review the critical technical components involved in constructing conversational search systems and explore their real-world applications. We will analyze query reformulation, clarification, conversational retrieval, response generation, and domain-specific applications to understand how these modules collaborate to create more natural and intelligent user interactions. Besides, we will summarize available resources and evaluation protocols to facilitate further research in this field. Our goal is to provide researchers and engineers with a comprehensive overview of the challenges and opportunities in conversational search. The paper will present the technical challenges at each stage, explore current solutions, and outline future directions to encourage further innovation and development in this emerging area.

Our paper differentiates itself from prior surveys on conversational search~\cite{book23_CIS_survey, book23_Neural_CIR_survey,DBLP:journals/csur/KeyvanH23,DBLP:journals/corr/abs-2407-00997} in two significant ways. First, earlier surveys were conducted prior to the emergence of ChatGPT\footnote{ChatGPT: \url{https://openai.com/chatgpt/}} and primarily focused on traditional methods without LLMs. LLMs have revolutionized the techniques and application scenarios in this area, leading to the development of real commercial conversational search engines. Our survey incorporates the latest LLM techniques in conversational search and discusses the associated new challenges and opportunities. Second, our survey adopts a more systematic approach. We structure our review from the perspective of information flow in conversational search engines, focusing specifically on the four key modules (i.e., query reformulation, clarification, conversational retrieval, and conversational response generation), while consistently discussing the integrative role and impact of LLMs throughout. A broad overview of conversational search is presented in Fig.~\ref{fig:taxonomy}, which categorizes existing studies using a structured taxonomy.

The remaining part of this survey is organized as follows: 
Section~\ref{sec: Query Reformulation},~\ref{sec:clari}, ~\ref{sec:retrieval}, ~\ref{sec: generation} review the progress and key issues from the perspectives of query reformulation, clarification, conversational retrieval, and conversational response generation, which are four key components of a conversational search system.
Then, Section~\ref{sec:domain} presents pioneering conversational search studies conducted in specific domains and user-centric scenarios. 
Section~\ref{sec:evaluation} summarizes the available benchmarks and discusses the evaluation protocols for improving conversational search systems.
Finally, we conclude the survey and discuss several potential future directions in Section~\ref{sec:conclu}.

\section{Query Reformulation} 
\label{sec: Query Reformulation}

Query reformulation is a widely used technique in the context of information retrieval~\cite{sparck2000probabilistic,gupta2023recent}, which aims to improve the search results by reformulating the initial user search query. 
This is because the input queries are usually vague, ambiguous, or incomplete, which requires the system to reveal the user search intent to address the information needs better.
Based on various reformulation techniques, such as query expansion, query rewriting, and term substitution, the search system can achieve more relevant and accurate results~\cite{carpineto2001information, nogueira2017task}.

\subsection{Query Reformulation in Conversational Search}
\label{sec: CQR model}

\begin{figure}[tp]
\centering
\includegraphics[width=.8\linewidth]{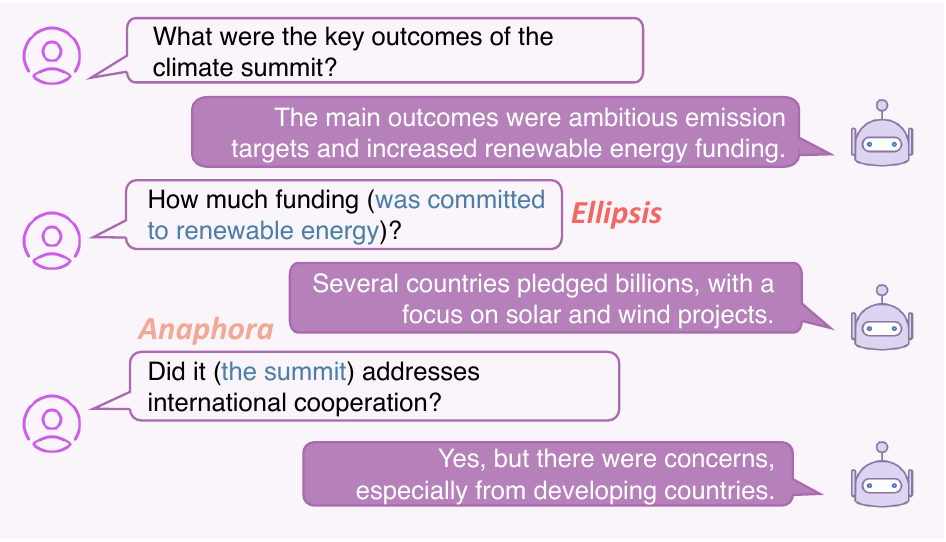}
\caption{An example illustrating the semantic phenomena of \textit{Anaphora} and \textit{Ellipsis} in conversational search. Query reformulation is crucial in addressing these semantic gaps.} 
\label{fig:qr}
\end{figure}

In conversational search, query reformulation is critical due to the complex nature of user intent, which is often obscured within follow-up questions and results in context-dependent queries~\cite{dalton2020cast,dalton2021cast,owoicho2022trec}. Unlike traditional information retrieval, conversational search must address two key semantic phenomena: \textit{Anaphora} and \textit{Ellipsis}~\cite{transformerplus,hongjin22}.
\textit{Anaphora} involves expressions of which semantics rely on previous context, while \textit{Ellipsis} refers to the omission of semantic elements understood from prior context. These phenomena necessitate that current queries are semantically linked to previous historical context, e.g., query-answer pairs, making context tracking essential for accurate responses. Fig.~\ref{fig:qr} shows a case of query reformulation in conversational search. Adaptive techniques~\cite{elgohary2019can} are required to resolve references and fill in omitted information based on conversation history. Consequently, effective query reformulation must be dynamic and context-aware, continuously adapting to the evolving discourse. 
From the literature, the existing methods can be divided into three categories: 1) Query expansion, 2) Query rewriting, and 3) Hybrid, which are introduced as follows.

\subsubsection{Query Expansion}
Due to the lack of query reformulation annotation, earlier works borrow the idea of query expansion from traditional IR~\cite{efthimiadis1996query} to expand the current query turn from different resources.
Various alternatives can be used to achieve query expansion, including relevance term classification, heuristic techniques, etc.

Relevance term classification transforms query reformulation as a binary classification task where terms in the query context are labeled as relevant or irrelevant. The relevant terms are then appended to the original query as expansion. This approach systematically incorporates important terms, thereby improving the retrieval performance.
Typically, Kumar et al.~\cite{kumar2020making} develop a binary term classifier with weak-supervision signals to select relevant terms from the dialog history and pseudo relevance feedback (PRF) to de-contextualize the conversational query. Voskarides et al.~\cite{voskarides2020query} propose QuReTeC with distant supervision, whose pseudo labels are readily available and can be inferred from user-system interactions. Lin et al.~\cite{lin2021contextualized,lin2021multi} estimate term importance during query reformulation, which evaluates the significance of different terms within the context, ensuring that the most critical terms are retained or emphasized in the reformulated query.
Besides, heuristic methods focus on leveraging linguistic features to reformulate queries, which leverage techniques like part-of-speech tagging, dependency parsing, and co-reference resolution to reconstruct the original query by identifying grammatical structures and resolving references to the previous context.
By utilizing syntactic and semantic analysis, these methods~\cite{mele2020topic,voskarides2020query} enhance the clarity and relevance with the context of the reformulated query and thus improve the search results.

Overall, the query expansion approaches achieve significant improvement in conversational search without relying on human labeling for model training. However, a remaining challenge is to distinguish the usefulness of the expansion terms~\cite{cao2008selectinggood,he2009cikm}, i.e., the pseudo supervision signals generated by the probability of language modeling or heuristic rules might not guarantee the effectiveness for search. To mitigate such issues, Mo et al.~\cite{ConvRelExpand} view the conversational query reformulation as a turn-level expansion task based on the pseudo labels generated by the direct search impact, selecting the relevant turns from the history to expand the original query. More explorations are desired with the help of existing large language models (LLMs), e.g., expanding the query from the generation results of LLMs.

\subsubsection{Query Rewriting}
Query rewriting is a fundamental approach to transform context-dependent queries into stand-alone ones conditioned on the previous conversational context, which has been applied in early studies and continues to evolve with advancements in LLMs.

In early studies, Yu et al.~\cite{yu2020few} propose a rule-based query simplification method that generates distantly supervised data to train a query generation model. Lin et al.~\cite{lin2020conversational} deploy sequence-to-sequence architecture models as a de facto principle to mimic the human-rewritten query and several following studies~\cite{transformerplus,del2021question,vakulenko2021comparison} further improve the query rewriting results. 
However, these approaches highly rely on the quality of human annotations to train the rewriter models, which might limit the upper bound of the conversational search performance.
Wu et al.~\cite{wu2022conqrr} first prove that the human-annotated rewritten query might not necessarily be the best search query. To this end, some works~\cite{wu2022conqrr,chen2022RLCQR} adapt reinforcement learning by using retrieval improvements as reward signals to optimize the rewriting model training toward downstream search tasks directly. Besides, the other studies~\cite{hongjin22,mao2023search} leverage edit-based methods to perform explicit query rewriting by injecting the ranking signals.
Inspired by the strong
capabilities for long-context understanding and text generation of LLMs, Mao et al.~\cite{mao2023large} first leverage LLMs as search intent interpreters for the first time to help conversational search and enhance the performance by fusing multiple prompting results. Ye et al.~\cite{ye2023enhancing} achieve informative query rewriting by multi-stage generation. To avoid the high cost by directly calling LLMs to conduct query rewriting and human annotated labor, Jang et al.~\cite{jang2023itercqr} initialize the training objective for each query turn by ChatGPT and design an iterative mechanism to improve the generated query according to the ranking results. 

Overall, the query rewriting technique is beneficial for solving the ambiguous nature of the query used in conversation search. The main challenge lies in optimizing the query rewriter model toward search results and aligning with a retriever for natural language-based queries.

\subsubsection{Hybrid Approaches}
The query expansion and query rewriting approaches produce different effects in terms of query reformulation under conversational scenarios. The query expansion aims to add supplementary information to the query, while the rewriting tends to deal with ambiguous queries and add missing tokens. Both effects are important and they are usually studied separately. To inherit their benefits, Mo et al.~\cite{mo2023convgqr} propose ConvGQR, a new framework for conversational query reformulation by combining the rewriting results and the generated answer as expansion from language models. The assumption is the stored knowledge enables the language models to generate semantic similar answers that might co-occur with the ground-truth in the relevant documents and thus facilitate the search procedure. This principle has been widely used in the era of LLMs~\cite{wang2023query2doc}. Since the LLMs obtain much more powerful question-answering ability with more parametric knowledge, the generated answers become important expansion parts together used with the rewritten queries to form the final query~\cite{mao2023large,yoon2024ask}. Besides, Mo et al.~\cite{mo2024chiq} design a two-step method that leverages different capabilities of LLMs to enhance the conversational history for conducting query reformulation, including but not limited to rewriting and expansion. With the development of various techniques in the field of IR and NLP, more sophisticated hybrid approaches are expected to enhance query reformulation.

\subsection{Analysis of Existing Datasets}
\label{sec: QR Analysis}
The development of conversational search datasets significantly facilitates the technique of query reformulation. 
Several dedicated datasets offer unique contributions from various aspects.
In this section, we analyze the existing datasets with the utilization of query reformulation, which aims to help the community develop and refine query reformulation models according to their construction details, e.g., benefits and challenges.

Typical query reformulation datasets designed for conversational search include CANARD~\cite{elgohary2019can}, QReCC~\cite{anantha2021open}, the TREC Conversational Assistance Track (CAsT) series~\cite{dalton2020cast,dalton2021cast,dalton2022cast,owoicho2022trec} etc. Their construction involves both manual annotation and automatic generation.
As aforementioned discussion, the quality of the annotation would greatly affect the search performance. 
Though with high cost, the human annotators manually rewrite the original queries to provide clear and explicit context, ensuring high-quality annotation.
This is done in the TREC Conversational Assistance Track (CAsT) and the recently improved TREC Interactive Knowledge Assistance Track (iKAT)~\cite{aliannejadi2024trec,abbasiantaeb2025conversational} series across the test set, and some of them are exemplified by the CANARD dataset in training data.
The rewritten queries provided in the QReCC dataset are inherited from CANARD and expanded automatically by replacing the pronouns with anaphora resolution rules. As a result, its linguistic phenomenon is not as abundant as the fully manually constructed datasets.
Besides, the relevance judgments for the documents associated with the original question are determined by whether containing the answers from the annotators, which might not guarantee the corresponding rewritten queries are optimal search queries. 
Thus, the discrepancy between document assessment and rewritten query annotation for each query turn should be carefully addressed when designing a query reformulation model for downstream search tasks.

\subsection{Evaluation of Query Reformulation}
\label{sec: QR assess}
The evaluation of query reformulation quality is critical to ensure the effectiveness of conversational search systems. Two primary methods are employed to evaluate the performance of reformulated queries: 1) lexical overlapping metrics and 2) end-to-end retrieval performance.

\noindent \textbf{Lexical Overlapping Metrics.}
The lexical overlapping metrics measure the accuracy of the reformulated query by comparing it to a reference query using token-level precision, recall, and F1-score. Precision evaluates the proportion of correctly generated tokens among all generated tokens, recall measures the proportion of correctly generated tokens among all relevant tokens, and the F1-score provides a harmonic mean of precision and recall. These metrics provide a quantitative assessment of how closely the reformulated query matches the reference query, offering insights into the exactness and completeness of the query rewriting process~\cite{elgohary2019can,anantha2021open}.

\noindent \textbf{End-to-End Assessment.}
Another method for evaluating query reformulation is end-to-end assessment, which measures the effectiveness of the reformulated query in final retrieval performance.
This principle can be applied to both sparse and dense retrieval systems. 
Similar to the ad-hoc search, after retrieving relevant documents by using the reformulated queries, their effectiveness is measured based on the retrieval metrics such as Mean Reciprocal Rank (MRR), Normalized Discounted Cumulative Gain (NDCG), and Recall at various cut-off levels (e.g., Recall@10, Recall@100). This end-to-end evaluation provides a comprehensive assessment of how well the reformulated queries perform in real retrieval scenarios, reflecting the practical impact of the query reformulation on the search results~\cite{ConvRelExpand,mao2023search}.

By employing both lexical overlapping metrics and end-to-end assessment methods, researchers can obtain a holistic understanding of the effectiveness of query reformulation techniques. This dual approach ensures that the reformulated queries not only align well with reference queries but also improve retrieval performance in practical applications.

\subsection{Limitation and Discussion}
Despite the advancements in conversational query reformulation techniques, several limitations could be addressed to further improve the robustness and effectiveness. 

\noindent \textbf{Dataset Bias.} Most existing datasets for query reformulation are automatically constructed, 
and even the small portion with human annotations, the ``golden queries'' may still be sub-optimal due to annotation bias. 
Such an issue arises from subjective interpretations and inconsistencies in labeling, which would result in sub-optimal training data. Furthermore, the limited size of the datasets restricts the ability of trained models to generalize and guarantee good performance across diverse conversational contexts, leading to unreliable query reformulation in practical applications. The development of larger and more diverse datasets with consistent and accurate annotations is desirable.

\noindent \textbf{Evaluation Challenges.} 
The lexical overlapping metrics for query reformulation assessment measure token-level precision, recall, and F1-score, which provide only an indirect evaluation. 
Because they cannot reflect the real effectiveness of the reformulated queries in terms of downstream search tasks.
The end-to-end assessments, although more comprehensive, might be influenced by model biases, such as they might be sensitive to various query encoders. These biases can obscure the actual performance of the query reformulation techniques, leading to the same issue as lexical overlapping metrics. Thus, the key point of the assessment is to ensure the connection between query reformulation techniques and the direct downstream search task performance.

\noindent \textbf{Challenge in Long Conversation.} 
In real-world scenarios with long dialogues, query reformulation errors might accumulate from each turn as the conversation goes on. This phenomenon becomes severe in longer conversations, which leads to a degradation in the overall performance of the conversational search system.
Besides, since the existing query reformulation techniques only focus on addressing the semantic phenomena, e.g., co-reference and omission in one natural language sentence, and expand the key terms from the context, which might not be enough for a good search query or injecting noise as the conversation diving in.
Different from query reformulation for ad-hoc search, the conversational scenario should consider a more appropriate mechanism in the future to identify useful information and omit the noisy ones from the context when reformulating the search queries.

\noindent \textbf{Future Paradigm of Query Reformulation.} 
The goal of query reformulation in existing retrieval systems, either ad-hoc or conversational search, is to supply missing search aspects or address ambiguity in the original query. 
With the development of large language models, some new paradigms of query reformulation have emerged, such as collaboration with different search agents~\cite{gong2024cosearchagent}, proactively providing reformulated query~\cite{avula2023and}, and reformulating queries with reasoning chains and generative language models~\cite{wang2023generative,zhang2024usimagent,engelmann2024context}. These studies improve query reformulation with various aspects and target different goals without a well-defined paradigm.
Furthermore, LLMs might implicitly reformulate queries during the generation process in any conversational systems, or advanced prompting techniques might directly elicit context-aware search behavior. Future work needs to explore the optimal interplay between explicit reformulation modules and the inherent contextual capabilities of LLMs, moving beyond optimizing reformulation in isolation.
Thus, it is still unclear what the destination of query reformulation is under the evolved conversational search paradigm, which requires further exploration from the community.

\section{Search Clarification} \label{sec:clari}

\begin{figure*}[t]
  \centering
  \includegraphics[width=\linewidth]{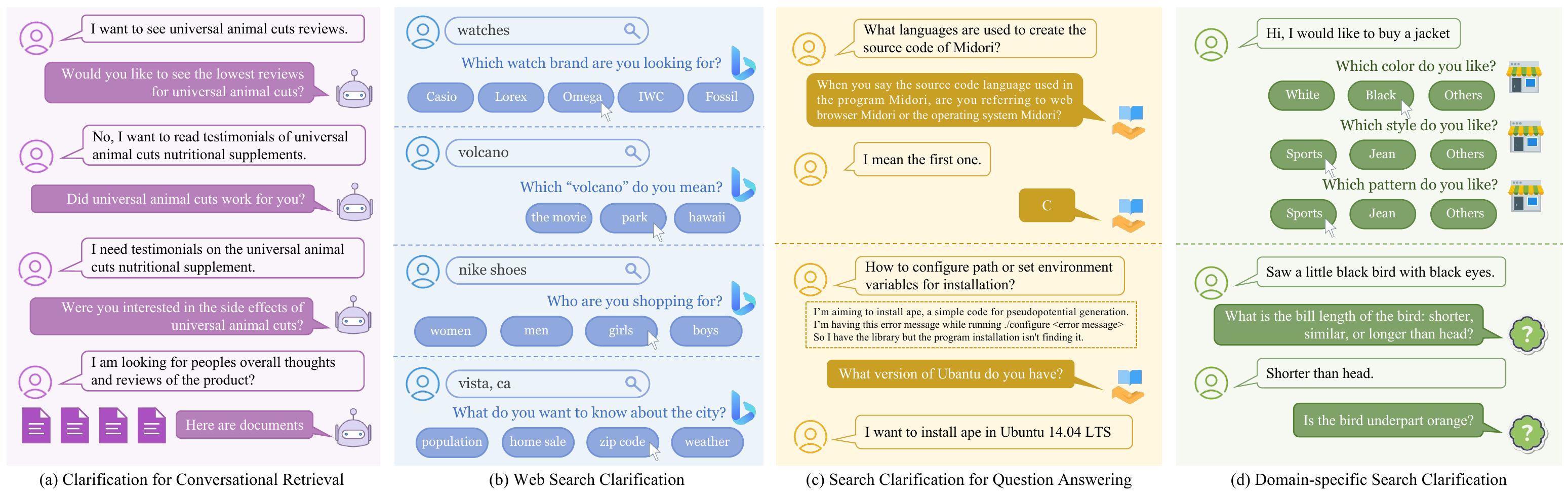}
  \caption{Different scenarios and examples of search clarification for conversational search systems, including (1) Clarification for conversational document retrieval, (2) Web search clarification, (3) Search clarification for Question Answering, and (4) Domain-specific search clarification.}

  \label{FIG_clr}
\end{figure*}

Conversational search provides users with an interactive interface enabling them to seek information, conduct behavior, or find a website with multi-turn interactions. During the conversation, users may deliver ambiguous, faceted, or unspecific search intents, which should be clarified by the systems for exact query understanding. To this end, the systems are expected to actively ask the users clarifying questions to help them re-articulate their intents. In this section, we summarize studies about asking clarifying questions in conversational search systems. Since the scenarios of different conversational search systems might be various, we first categorize the clarification scenarios as four main parts shown in Fig.~\ref{FIG_clr}, including (1) Clarification for conversational document retrieval, (2) Web search clarification, (3) Search clarification for question answering, and (4) Domain-specific search clarification. As for the domain-specific search clarification, we further divide it into different specific domains, including conversational recommender systems, legal search clarification, search clarification for programming, etc. We further summarize studies focusing on search clarification with large language models (LLMs) and finally discuss future directions.

This taxonomy categorization for search clarification scenario elucidates current technical boundaries and identifies promising pathways for developing context-aware clarification architectures, where each of them has distinct objectives, necessitating specialized datasets and learning paradigms. 
In conversational document retrieval clarification, which requires identifying relevant documents from a corpus, user queries are typically formulated as natural language questions. 
The primary challenge lies in generating clarifying questions that effectively disambiguate document clusters through strategic partitioning. 
In contrast, web search clarification focuses on addressing the inherent ambiguity of short queries commonly encountered in commercial search engines, where the challenges involve resolving semantic under-specification within limited lexical context.
In the context of question answering systems, which operate in knowledge-intensive environments, the search clarification is crucial for precise factual resolution. 
Besides, domain-specific search clarification introduces unique complexities from specialized terminologies and vertical knowledge requirements. The inherent differences in operational goals and contextual limitations pose substantial challenges to developing a cohesive framework for cross-scenario clarification systems.

\subsection{Clarification for Conversational Retrieval} \label{sec:clari-cs}

Conversational retrieval is a crucial part of a conversational search system. In the retrieval process, search clarification is defined as the multi-turn interaction between the user and the system~\cite{aliannejadi2019asking} as shown in Fig.~\ref{FIG_clr} (a). 
Formally, in each interaction turn $k$, the user inputs a query ($k = 1$) or responds to the clarifying question returned by the system ($k > 1$). For the system, it asks a clarifying question and retrieves relevant documents using obtained contexts simultaneously in each turn. 
The fundamental idea of a conversational retrieval system with clarification is to guide users through multiple rounds of queries, prompting them to provide increasingly useful information so the system can return more relevant documents in line with the user's intentions~\cite{krasakis2020analysing}.
In this section, we first summarize the available datasets for clarification in conversational retrieval and introduce clarifying question selection and generation methods. 
Then we further explore analytical studies that focus on data features instead of proposing new methods. Finally, we introduce some other relevant studies to improve the performance of the systems.

\subsubsection{Available Datasets}

Several datasets for conversational retrieval clarification are available to support methodology development.
\textit{Qulac}~\cite{aliannejadi2019asking} and \textit{ClariQ}~\cite{aliannejadi2020convai3} are two commonly used datasets that are built on top of the TREC Web Track 2009-2012 data~\cite{DBLP:conf/trec/ClarkeCV12}. Since these two datasets only contain clarifying questions and answers written by humans by crowd sourcing which are not applicable for many scenarios e.g., training a generation model,~\citet{aliannejadi2021building} further collect synthetic conversations to enhance these two and benchmarks several state-of-the-art methods for search clarification. They also propose a pipeline for online and offline evaluation.
~\citet{sekulic2021towards} propose another incremental dataset \textit{ClariQ-FKw} based on ClariQ. They extracted facet keywords in clarifying questions as the key information. Recently,~\citet{yuan2024asking} release the first multi-modal clarifying question dataset \textit{Melon} which extends existing study scenarios by incorporating multi-modal information. The construction of the datasets for search clarification in conversational retrieval is crucial and makes the development of the systems possible.

\subsubsection{Clarifying Question Selection} \label{sec:clari-csqs}

Most early studies based on Qulac and ClariQ datasets usually focus on how to select a clarifying question at each turn from a fixed-size question pool to reveal the user's search intent, because the datasets only support exploring within a simple setting.
Among them, ~\citet{aliannejadi2019asking} first design a framework consisting of clarifying question retrieval and selection, and document retrieval. Besides,~\citet{hashemi2020guided} propose Guided Transformer to help select the optimal clarifying question by leveraging external retrieved documents.
~\citet{bi2021asking} further apply negative feedback to generate yes/no-formed questions to explore the user's search intent within the several last conversation turns.
~\cite{mass2022conversational} point out that the retrieved documents play an important role in selecting the optimal clarifying question at each turn and they develop two fine-tuned models for question retrieval and ranking.
Although these approaches can achieve clarification needs, the size of the candidate clarifying question limits the capacity of the system response.

\subsubsection{Clarifying Question Generation} \label{sec:clari-csqg}

Selecting clarifying questions from a fixed-size question pool cannot satisfy the users' complex search intents in real-world scenarios.
Thus, a much more preferable paradigm is to generate a clarifying question based on the complex actual needs. 
To this end,
~\citet{sekulic2021towards} first propose a facet-based method to fine-tune GPT-2 to generate clarifying questions for conversational retrieval. They extract keywords from questions to train a facet prediction model, and then utilize the predicted facets to guide a GPT-2 model to generate questions. Recently,~\citet{wang2023zero} study how to generate clarifying questions in a zero-shot setting by prompting the language model with a facet-constrained prefix, which reduces the annotation cost.

\subsubsection{Analytical Studies}

Search clarification is an important component to achieve mixed-initiative interactions between the system and the user.
To provide a comprehensive understanding of how search clarification contributes to conversational search systems, some analytical studies are conducted.
~\citet{krasakis2020analysing} first analyze whether clarifying questions and user answers can influence the document ranking ability in conversational retrieval systems and whether the question or the answer has a more significant impact on the ranking results. Furthermore, ~\citet{aliannejadi2021analysing} analyze how the order of the feedback and the retrieval strategies influence the user's search experience. They reveal that query clarification acts better when the system asks first, whereas query suggestions are better when the system asks after search results are presented. Recently,~\citet{sekulic2024estimating} further propose a classification algorithm to measure the usefulness of the clarifying question and the user's answer.
These analytical studies empirically show how effective to implement a search clarification mechanism within a conversational search system.

\subsubsection{Information Integration} \label{sec:clari-cses}

Although clarifying questions can be generated, it is hard to collect the corresponding response for the question delivered by the user to either improve or evaluate the systems. 
Thus, a common practice is to simulate the response/feedback of the user and integrate them to improve the performance of clarification or search.
To this end,~\citet{wang2024depth} propose a method for simulating the user response and investigating how users reply to clarifying questions and their impacts.
Then, the system is verified that can better provide clarifying questions and retrieve search results by the designed simulators.
Besides,~\citet{yuan2024asking} argue that non-textual information (e.g., image and video) can provide additional information for search clarification. They construct a dataset Melon and propose an effective pipeline for multi-modal search clarification. 
More effective ways to automatically generate and integrate additional information, e.g., user simulation, are desirable in future exploration.

\subsection{Web Search Clarification} \label{sec:clari-web}

The aforementioned studies for search clarification in conversational retrieval are limited by only selecting the clarifying questions and evaluating the response in a fixed set.
Different from them,~\citet{zamani2020generating} propose clarification in web search,
In web search, the form of the user's query is more flexible, which can be a conversational query or a keyword query. When the user's query is ambiguous or faceted, the system can ask the user a clarification question, and provide multiple candidate facets for the user to choose from, which form a clarification pane as shown in Fig.~\ref{FIG_clr} (b). 
In the clarification pane, the clarification question is usually generated to improve the user's search experience, and the candidate facets are a set of generated or selected terms representing a group of high-coherency user sub-intents that the user may be interested in. When the user clicks one of the given facets, the query will be reconstructed by concatenating with the selected facet to form a new query and retrieve a new list of web pages~\cite{zhao2022generating}.

\subsubsection{Datasets}
There are several datasets are available to support the exploration in web search clarification.
Among them, the \textit{MIMICS} is the first dataset proposed by~\cite{zamani2020mimics}, which is based on the search log of the Bing search engine, so it contains rich information about user interests. The MIMICS dataset is divided into three subsets, namely MIMICS-Click (\~400k), MIMICS-ClickExplore (\~200k), and MIMICS-Manual (\~2k). The first two are completely based on Bing search logs, with a large number of queries and online user feedback, and are usually used as training data. The difference between the two is that each query in MIMICS-ClickExplore corresponds to multiple possible clarification panes, while each query in MIMICS-Click only contains a unique clarification pane. The queries in MIMICS-Manual contain manual quality annotations and are usually used for evaluation. After that,~\citet{tavakoli2022mimics} further release another dataset \textit{MIMICS-Duo} for multi-dimensional offline and online evaluation of web search clarification.

\subsubsection{Generating Clarifying Questions} \label{sec:clari-webcq}

In web search clarification, the crucial question is how to deliver a natural and informative clarifying question to the user.
~\citet{zamani2020analyzing} show that a good clarifying question can improve the search experience by grasping the intelligence of the systems. To generate good clarifying questions,~\citet{zamani2020generating} made the first step to propose three algorithms based on question templates and search logs, which laid the foundation for subsequent studies. They further propose a Transformer-based framework~\cite{zamani2020analyzing} for selecting the optimal clarification pane. After that,~\citet{wang2021template} propose a novel template-based method to let the model fill the slot in the question template to form a clarifying question.
However,~\citet{zhao2022generating} argue that template-based methods are difficult to find the descriptions of the query and facets, so they propose to find the descriptions from the web search results of the query. After that,~\citet{zhao2023improving} design a BART-based seq2seq model to generate clarifying questions and candidate facets simultaneously. Recently,~\citet{rahmani2024clarifying} investigate the usefulness of clarification by analyzing data in MIMICS~\cite{zamani2020mimics} and MIMICS-Duo~\cite{tavakoli2022mimics}. They conclude that specific, positively sentimental-oriented, lengthy, and subjective questions are useful for improving user experiences. The above studies analyze the role of clarifying questions from multiple perspectives and propose corresponding generation methods.

\subsubsection{Generating Candidate Facets} \label{sec:clari-webcf}

Unlike search clarification in other scenarios, web search clarification not only asks a clarifying question but also provides users with several candidate facets to click on. This facilitates user interaction and is more suitable for open-domain scenarios. Before this, query facets is a common query intent mining solution, aiming to mine comprehensive, multi-dimensional query sub-intents. For example, there are some rule-based and machine learning-based methods to mining these terms from the search result pages of the query~\cite{dou2011finding, kong2013extracting, kong2014extending,lu2025zero}.

Different from these studies, web search clarification focuses on finding a set of facets that users are most interested in, rather than showing all of them to users, which helps to concentrate on users' interests with providing better experiences. 
Following this idea,~\citet{hashemi2021learning, hashemi2022stochastic} first propose generating these facets using a seq2seq framework.
~\citet{samarinas2022revisiting} summarize all rule-based, machine learning-based, and pre-trained language model-based methods to generate the candidate facets for search clarification and conclude that BART-based models taking the query and search result snippets as the input show the best performance in almost all evaluation metrics.~\citet{zhao2023improving} deem that candidate facets are usually in the form of lists, so they collect parallel structures from the search result pages and use this information to enhance the generation ability of the BART-based model.~\citet{litvinov2024analyzing} analyze the whether and how the coherency of candidate facets influence the generation performance and proposed a coherency-based method for generation.~\citet{zhao2024generating} think that direct generation of candidate facets is not suitable for multi-turn usage, so they propose a framework to generate candidate facets within a multi-turn manner.

Recently, with the development of LLMs, some LLM-based methods have been proposed for generating candidate facets for web search clarification. Among them,~\citet{ni2023comparative} conduct a comparative study to compare different training objectives and base models for generating clarification facets.~\citet{lee2024enhanced} fully rely on LLMs to generate the facets. We believe that facets generation based on LLMs will become a more effective approach because LLMs contain much general and domain-specific knowledge.

\subsubsection{Other Related Studies} \label{sec:clari-webes}

Existing web search clarification studies usually generate several sub-intents as clarification candidate options. However, in some cases, users are also interested in parallel intent movement. 
To this end,~\citet{liu2024mining} design three algorithms to mine the parallel movement facets from search results and define the facets as exploratory queries. Except for generating clarification panes,~\citet{sekulic2021user} learn how to predict user engagement feedback using the data collected from a real-world search engine, and~\citet{lotze2021ranking} study ranking clarifying question candidates using these predicted user engagements. Furthermore,~\citet{gao2022search} utilize these user engagements prediction information and design a query-intent-clarification graph attention mechanism to select the optimal clarification pane.

\subsection{Search Clarification for Question Answering} \label{sec:clari-qa}

In question answering (QA), the goal is to answer the user's specific question.
However, the question can contain ambiguity, especially under a conversational scenario. Thus, it is necessary to ask a question for disambiguation. In this section, we classify existing studies in search clarification for QA as (1) Community Question Answering and (2) Knowledge-based Question Answering. As shown in Fig.~\ref{FIG_clr} (c), for Community QA, the system usually focuses on ambiguous intents in a domain-specific scenario such as programming, while the clarification usually focuses on some entity ambiguities in the knowledge-based QA.

\subsubsection{Datasets}

In some community QA studies~\cite{rao2018learning, rao2019answer}, the data is extracted from some community QA websites such as \textit{StackExchange} and \textit{Amazon}. These websites contain lots of ambiguous questions, user answers, and clarifying questions. Later,~\citet{kumar2020clarq} propose a dataset \textit{ClarQ} based on StackExchange. To our best knowledge, the scale of ClarQ is much larger than other datasets based on StackExchange. On the other hand, in knowledge-based question answering, \textit{Abg-CoQA}~\cite{guo2021abg} is proposed to tackle ambiguity in conversational QA, which focuses on disambiguation for human-to-human conversations.

\subsubsection{Community Question Answering} \label{sec:clari-qacqa}
Community Question Answering (Community QA) aims for a collaborative online platform where users can ask questions and receive answers from other community members. Building up such platforms relies on collective knowledge beyond a single expert or automated system. 
In community QA, several users might engage in multiple turns of discussion around a certain topic, thus clarification interactions frequently occur~\cite{rao2018learning}. Therefore, researching how to automatically clarify user questions is crucial.
To this end, \citet{braslavski2017you} make the first step to analyze clarifying questions in community QA. They study the user questions from two domains and analyze the user behavior and the taxonomy of the asked clarification questions. After that, Rao et al.~\cite{rao2018learning, rao2019answer} design different methods to generate good clarifying questions for community QA. \citet{trienes2019identifying} is the first study investigating whether a community QA question is clear or not. They propose a method of judging whether a question needs further clarification.~\citet{kumar2020ranking} study how to rank clarifying question candidates by natural language inference. As the clarification mechanism is complicated in community QA, some studies~\cite{tavakoli2020generating, tavakoli2022analyzing} made some analytical studies about the feature of good clarifying questions, the taxonomy of clarification questions, and the types of user answers. Although these studies have proposed effective methods to generate community QA questions, these methods are difficult to transfer to other open-domain scenarios. 

\subsubsection{Knowledge-based Question Answering} \label{sec:clari-kqa}

In knowledge-based question answering systems, the clarification process usually targets reference issues between entities.~\citet{xu2019asking} propose a dataset \textit{CLAQUA} focus on solving clarification in knowledge-based question answering. They also figure out single-turn and multi-turn ambiguity and design to tackle these two situations.~\citet{nakano2022pseudo} explore the impact of generating pseudo-ambiguous and clarifying questions based on sentence structure on the clarification question answering system. This study provides useful guidance for optimizing the clarification question answering system, enabling it to better understand user needs and generate more targeted clarification questions.~\citet{kim2023tree} study using retrieval-augmented large language models to handle ambiguous questions.~\citet{lee2023asking} provide an in-depth analysis for understanding how to interact with users in open-domain question answering systems and how to adjust their behavior according to user needs. These studies are important for understanding users' ambiguous intent in conversational search by solving the entity ambiguity problem.

\subsection{Domain-specific Search Clarification} \label{sec:clari-domain}

Aside from the aforementioned general search clarification scenarios, some studies focus on a closed search scenario, such as Conversational Recommender System, Legal Search, etc, which are also important. Fig.~\ref{FIG_clr} (d) illustrates two close-domain settings, which require the methods to be adapted accordingly to the scenarios. We introduce and summarize related studies for different domain-specific search clarification scenarios in this section.

\subsubsection{Conversational Recommender System} \label{sec:clari-crs}

The Conversational Recommendation System (CRS) is also a search clarification scenario where users need to buy products, and the system gradually narrows the range of products by asking questions about different attributes of the products, making it easier for users to find the products they want to buy. We list some representative related studies in recent years~\cite{christakopoulou2016towards, li2018towards, sun2018conversational, chen2019towards, zhou2020towards, zhou2020improving, lei2020interactive, lei2020estimation, deng2021unified, zhang2022multiple, friedman2023leveraging, wang2023rethinking}.
Since CRS can be regarded as an independent research field, there are already some reviews~\cite{jannach2021survey, gao2021advances, lei2020conversational} that comprehensively summarize the related work with more details.

\subsubsection{Legal Search Clarification} \label{sec: SC-legal}

In legal search, users often deliver ambiguous intents due to the complexity of legal documents. Therefore, a clarifying question can be effective in helping the user articulate her complicated search intent.~\citet{liu2021conversational} compare the search behavior and outcomes of conversational and traditional search in legal case retrieval. After that, they further explore how buffer mechanisms and generated queries can help improve the performance of conversational agents in legal case retrieval~\cite{liu2022query}. Furthermore, they also try to study the behavior of conversational agents for retrieving better legal cases. This study provided an in-depth analysis for understanding how conversational agents interact with users and how to adjust their behavior according to user needs. Recently,~\citet{liu2023leveraging} introduce event schema, where conversational agents can have a better understanding of the user's potential intent and ask targeted clarifying questions to improve the accuracy and efficiency of retrieving legal cases.

\subsubsection{Search Clarification for Other Scenarios} \label{sec:clari-other}

Search clarification can further be applied in any scenario where the user's query or question can be ambiguous or faceted. In this section, we list some other scenarios of search clarification as follows which can be used to expand clarification to new scenarios and can provide inspiration: interactive classification~\cite{zamani2020generating}, image guessing~\cite{white2021open}, product description~\cite{majumder2021ask, zhang2021diverse}, collaborative building~\cite{shi2022learning}, programming~\cite{eberhart2022generating, li2023python}, user simulation~\cite{erbacher2022interactive}, social or moral situation~\cite{pyatkin2023clarifydelphi}, task-oriented dialogues~\cite{feng2023towards}, math~\cite{mansouri2023clarifying}, spatial reasoning~\cite{deng2023learning}, and collaborative dialogue~\cite{testoni2024asking}.

\subsection{Search Clarification with LLMs} \label{sec:clari-llm}

Recently, with the development of large language models (LLMs), there are also some studies on how to let LLMs ask clarifying questions. For example, in clarification for conversational retrieval, LLMs can be used to generate clarifying questions and simulate user responses~\cite{wang2023zero, wang2024depth}. For the web search clarification, the LLMs can be applied to generate better clarifying questions or query facets~\cite{ni2023comparative, lee2024enhanced, zhao2024generating, liu2024mining}. In conversational recommender systems, leveraging LLMs can also improve the recommendation effectiveness~\cite{friedman2023leveraging, wang2023rethinking}. These studies have shown that LLMs help existing scenarios to improve the system's performance due to LLMs' strong natural language generation and instruction-following capability.

With the powerful natural language generation capabilities of LLMs, the existing tasks and scenarios can be effectively improved. However, 
it is unclear whether LLMs can be used to expand the existing clarification scenarios so that clarification can have a wider definition and application range. To this end,
~\citet{mu2023clarifygpt} propose ClarifyGPT to ask clarifying questions for code generation. Some studies explore new search clarification scenarios. For example,~\cite{li2023eliciting} elicit user interest by asking clarifying questions generated by LLMs.~\citet{andukuri2024star} further design a reinforcement learning system to generate better clarifying questions.~\citet{deng2023rephrase} propose a ``Rephrase and Respond (RaR)'' framework to let the LLM to generate clarifying questions.

\subsection{Future Discussion}

As mentioned in this section, search clarification has been studied and applied in various types of conversational search systems, including (1) clarification for conversational (document) retrieval, (2) web search clarification, (3) search clarification for question answering, and (4) domain-specific search clarification. We also introduce search clarification with LLMs. However, although many existing scenarios have explored the effectiveness of search clarification, we can still have some future discussions on many aspects of the era of LLMs. In this section, we propose three potential future research directions for search clarification.

\noindent \textbf{Search Clarification for LLMs.} In Section~\ref{sec:clari-llm}, we summarize recent studies about search clarification with LLMs. These studies focus on utilizing the ability of LLMs as a tool for improving search clarification effectiveness in conversational search systems. However, in some cases, the LLM itself can also be seen as a conversational search system that has a large amount of user logs. Thus, search clarification for LLMs themselves is also worth studying.
Besides, search clarification is crucial for understanding user input and might also be important for clarifying LLMs' output or behavior, e.g., explain why retrieve this document. As LLMs enable more fluid, integrated conversations, clarification might become an inserted dialogue management strategy, seamlessly integrated within the retrieval and generation flow to maintain mutual understanding and guide the interaction effectively.

\noindent \textbf{Search Clarification for Retrieval-Augmented Generation.} Some current hot search scenarios can also introduce clarification for enhancement, such as Retrieval-Augmented Generation (RAG). For example, when the user asks a question including an ambiguous entity, the system can ask a question to let the user disambiguate the query and then do the retrieval process. The system can also identify ambiguous or faceted intents embodied in the question to ask more natural clarifying questions.

\noindent \textbf{Search Clarification in Vertical Domains.} 
We can explore more natural clarification generation in some vertical fields, such as medical consultation, customer service robots, etc. The design could focus more on the semantic differences in different domains.

\section{Conversational Retrieval}
\label{sec:retrieval}

Given a multi-turn information-seeking conversation, 
a conversational search system aims to retrieve relevant passages from a large collection, which is considered to be more complicated than traditional ad-hoc retrieval due to the comprehension requirement for the conversational context and the current user query.
Besides, user-system conversations often contain rich language patterns and topic switches, which increase the difficulty of context modeling. The challenges that hinder the conversational retrieval models come with different aspects, as introduced as follows.

First, although recent neural models, e.g., BERT~\cite{bert}, T5~\cite{t5}, and recently emerged LLMs~\cite{gpt4}, have shown the advanced ability of sequence modeling, it is still a long-term challenge to fully interpret the multi-turn natural language form conversations. 
Second, the transitions of search intents will introduce much noise into the historical context, which is irrelevant to the current search behavior~\cite{coted,ConvRelExpand}.
Thus, it is crucial to address how to denoise the conversational context.
Third, the training data for conversational retrieval models is limited.
Labeling multi-turn interactions is time-consuming for human annotators, and the relevance signals of passages are difficult to gain.
Therefore, various levels of data augmentation techniques, such as generating queries, context-passage pairs, and whole conversations have been developed.
Lastly, though conversational retrieval models can usually achieve better performance than query reformulation approaches, the lack of interpretability hampers the understanding of dense representations in conversations.
To address these challenges, the existing studies of conversational retrieval can be categorized into four groups: (1) conversation modeling, (2) context denoising, (3) data augmentation, and (4) interpretation.

\subsection{Conversation Modeling}
\label{subsec:Conversation_Modeling}

Unlike traditional single-turn retrieval, modeling multi-turn conversations requires a more sophisticated model architecture design.
HAM~\cite{ham} is an early work that utilizes the pre-trained language model (PLM) BERT~\cite{bert} and the attention mechanism to perform soft selection for conversation histories.
This model assigns different weights to historical interactions based on their relevance and usefulness in resolving the current information needs.
It also adapts multi-task learning to perform the conversational response generation task and the dialog act prediction task to learn a more robust representation of the conversation.
Qu et al.~\cite{ORConvQA} proposes a standard approach for modeling conversational context by concatenating all historical queries within a conversation and uses a simple PLM-based encoder to model this concatenation into a dense representation for retrieval task.
Le et al.~\cite{CoSPLADE} utilizes the sparse retriever SPLADE~\cite{SPLADE} to model the conversations into the sparse representation of queries and passages and Jeong et al.~\cite{jeong2023phrase} modeling the conversational dependency to achieve phrase-level retrieval for search results.

Aiming to better understand the intricate correlations among user behaviors in conversations, some researchers leverage graph-based approaches to model the conversations to capture the connection between different query turns and the retrieval pool.
Li et al.~\cite{DHM} deploy a graph-guided approach of retrieval to model the internal dependency between the responses of different turns and the current user query.
Further, they develop a framework DGRCoQA~\cite{DGRCoQA} that comprises three components: a dynamic question interpreter, a graph reasoning enhanced retriever, and a standard reader, to construct a dynamic context graph to model the response flow of the conversation.

In the era of LLM, the powerful context understanding ability enables the community to design a more sophisticated approach to conversational modeling.
Pan et al.~\cite{Conv-CoA} develop a chain-of-actions for conversational retrieval, which devises a framework of systematic prompting, pre-designed actions, and a Hopfield-based retriever.
Mao et al.~\cite{ChatRetriever} use a simple but effective conversational instruction tuning technique to train an LLM based on contrastive learning, which performs enhanced complex conversation modeling for conversational retrieval.
This work explores the potential of aligning LLMs for retrieving relevant passages under multi-turn scenarios.

\begin{figure}[tp]
\centering
\includegraphics[width=.95\linewidth]{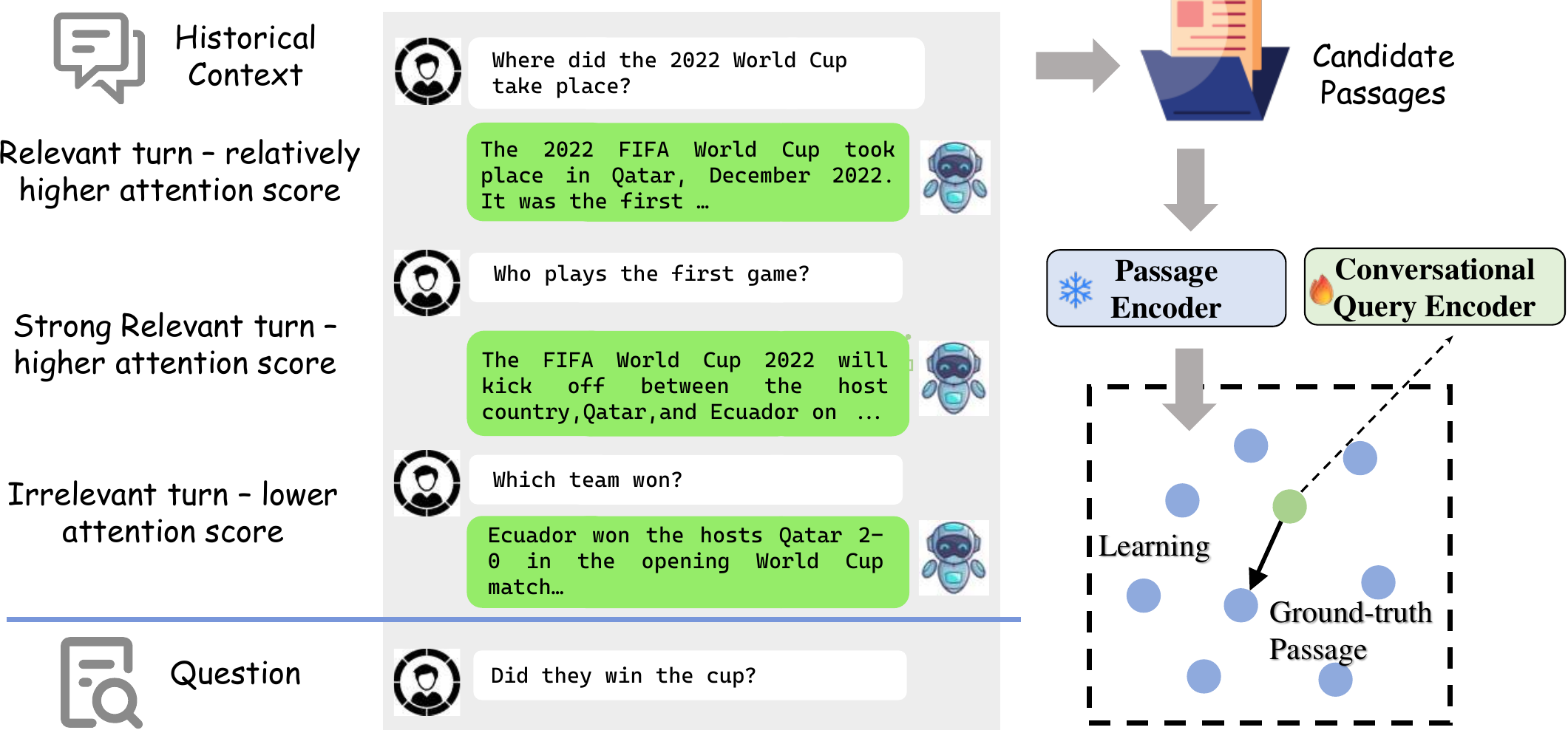}
\caption{An example of the end-to-end learning of the model for conversational dense retrieval with implicit de-noising. Each historical turn would have a certain degree of correlation with the current turn, e.g., irrelevant, partially relevant, or strongly relevant.} 
\label{fig:cdr}
\end{figure}

\subsection{Context Denoising}
\label{subsec:Context_Denoising}

Within a conversational session, not all historical query turns are relevant to address the current user's information needs. 
Simply concatenating all historical context would inject unnecessary and noisy turns and thus degrade the performance of the system. 
Thus, accurately identifying the useful historical context of the conversation is a crucial research question. 
To this end, context denoising techniques aim to mitigate the negative impact of irrelevant or noisy historical context, which can be categorized as implicit or explicit.
Implicit context denoising techniques achieve this goal by relying on model architectures (e.g., attention mechanisms) or training strategies (e.g., contrastive learning), which inherently learn to focus on relevant parts of the context. 
Then, the weight of irrelevant or noisy information would be reduced or ignored during representation learning without explicitly selecting or removing historical turns.
On the other hand, explicit context denoising methods explicitly select or remain relevant historical elements.

\subsubsection{Implicit Context Denoising}
\label{subsubsec:Implicit_Context_Denoising}
Some works attempt to perform context denoising implicitly, and a common practice is shown in Fig.~\ref{fig:cdr}, aiming to teach the model to implicitly identify to what extent (e.g., attention scores) a historical turn is relevant to the current query.
For example, Yu et al.~\cite{ConvDR} leverages dense retrieval to embed conversational queries and documents into a shared vector space where semantic similarity can be directly measured. 
It employs a teacher-student framework, which involves using an existing, well-trained dense retriever as a teacher to guide the student model, by inheriting its document encodings and learning to mimic the teacher's embeddings for queries. 
This strategy enhances model ability by capturing the necessary context while ignoring unrelated information to improve retrieval accuracy across conversational turns.
Mao et al.~\cite{coted} develop a framework by combining curriculum learning and contrastive learning to progressively guide a conversational query encoder on how to filter out irrelevant information from the conversation history. 
It involves generating noisy and clean versions of conversational data and using these to train the model by gradually increasing the complexity of the training data.
Krasakis et al.~\cite{zeco} aim to address context denoising by leveraging existing ad-hoc search data to understand and filter out noisy context, improving the relevance judgment of retrieved passages without additional fine-tuning on conversational datasets.
Specifically, it contextualizes the query embeddings within the conversation while ensuring that only the current query terms are matched with potential answers.
More recently, Mo et al.~\cite{QRACDR} propose QRACDR to first define a better aligned embedding space for the query representation learning and enable to model to learn to align the query representation with this area via rewritten queries and relevance judgments, which obtains a less noisy context representation implicitly. Then, Lupart et al.~\cite{lupart2025disco} follow this principle to leverage LLM to obtain further improvement.

\subsubsection{Explicit Context Denoising}
\label{subsubsec:Explicit_Context_Denoising}
Instead of implicitly conducting context denoising, some researchers attempt to identify and select useful historical conversational context explicitly.
For example, Mo et al.~\cite{ConvRelExpand} focus on selecting historical queries that are relevant and can enhance the current one, rather than using all previous queries, which may introduce noise and degrade search performance. They design a pseudo-labeling algorithm to identify the usefulness of the queries in history according to the impact on search results, subsequently training a selection model on these pseudo labels. 
A multi-task framework is designed to simultaneously optimize both the selector and the passage retriever, maintaining alignment between the selection of historical queries and the retrieval operations.
Following the previous observation, they further develop HAConvDR~\cite{HAConvDR} to enhance the capability of conversational search systems to handle complex user-system interactions, e.g., with frequent topic shift and long noisy context, which is based on a context-denoised query reformulation strategy that selectively incorporates relevant historical information into the current query and a mechanism to mines additional supervision signals from historical turns to guide the history-aware training process.
The context denoising mechanism significantly enhances the performance and robustness of the conversational search systems by adapting to practical scenarios.

\subsection{Data Augmentation}
\label{subsec:Data_Augmentation}

Effective query understanding within conversational retrieval systems relies heavily on large-scale datasets, due to the long-tail and complex linguistic phenomenon in multi-turn scenarios. However, the abundant annotations are usually unavailable to make the training procedure feasible.
To enrich the conversational retrieval data, various levels of data augmentation are explored in the literature, including 1) relevance judgment generation, 2) contrastive sample generation, and 3) conversation session generation.

\subsubsection{Relevance Judgment Generation}
\label{subsubsec:Relevance_Judgment_Generation}
In the context of information retrieval, manual relevance judgments are expensive to obtain, while such an issue becomes much more obvious in conversational search.
To mitigate the scarcity of annotation, a common practice in the literature is to generate them automatically.
For example, Lin et al.~\cite{CQE} create pseudo-relevance labels based on the pseudo relevance feedback to help guide the model to learn how to integrate conversational query reformulation directly into a dense retrieval framework via an end-to-end manner, which achieves a more coherent optimization across stages compared with separately.
Jin et al.~\cite{InstructoR} leverage LLMs to provide a form of supervision signal that can guide the training of a conversational retriever without the need for extensive manual labeling.
The principle is to deploy LLMs as an ``instructor'' to generate soft labels, which indicate the relevance of session-passage pairs through an unsupervised training framework with the relevance score estimated by the LLMs.
This approach sidesteps the data scarcity issue by harnessing the generative capabilities of LLMs for data augmentation in conversational retrieval tasks, which is a valuable direction for further exploration.

\subsubsection{Contrastive Sample Generation}
\label{subsubsec:Contrastive_Sample_Generation}
Sufficient supervision signals, either positive or negative, are important for dense retrieval~\cite{karpukhin2020dense} with contrastive learning~\cite{robinson2021contrastive}.
For conversation scenarios, additional contrastive samples can be generated from the conversational history.
The basic assumption~\cite{HAConvDR} is among the top-ranked historical retrieved results, the ones highly relevant to the current query can be used as pseudo positive samples, while others are less relevant and can serve as hard negatives for training.
Besides, Kim et al.~\cite{Shortcut} identify a ``retrieval shortcut'' phenomenon,
which occurs when models overly rely on partial conversation history, but ignore the current query and results in topic drift. 
To this end, \citet{HAConvDR} mines additional hard negatives from the historical turns to mitigate the dependency on retrieval shortcuts by forcing the model to consider both the current question and the entire conversation history. 

\subsubsection{Conversation Session Generation}
\label{subsubsec:Conversation_Session_Generation}
As search systems under conversational scenarios are not yet widely implemented, it is challenging to obtain enough high-quality session-level search data to facilitate the fine-tuning of conversational dense retrievers.
To address this, Mao et al.~\cite{ConvTrans} propose a data augmentation solution by transforming readily available web search sessions into conversational search sessions, which is based on the structural and behavioral similarities between web search sessions and conversational search sessions. 
It reorganizes raw web search sessions into a heterogeneous graph that represents the multi-turn nature of user interactions, which is then enriched with additional queries to improve diversity.
Then, a random walk sampling algorithm is employed to generate pseudo conversational search sessions that can be used to train conversational dense retrieval models.
With the development of LLM, Mo et al.~\cite{ConvSDG} leverage LLMs to create additional conversational session data to enhance the diversity and volume of training data through two approaches: dialogue-level session generation, which creates entire sessions based on a given topic, and query-level augmentation, which rephrases existing queries to express the same intent in various ways. 
Similarly, Huang et al.~\cite{huang2023converser} propose CONVERSER to provide guided examples to achieve few-shot generation.
Chen et al.~\cite{ConvAug} propose ConvAug, which is driven by the fact that the diversity of conversational expressions is limited in existing conversational search datasets and thus makes it difficult for models to adapt to real-world complexities.
Inspired by human cognition to improve the quality of generated data and reduce errors, ConvAug leverages LLMs to 
generate multi-level augmented conversations to capture a wider range of conversational contexts. 
Additionally, it employs a difficulty-adaptive sample filter to select challenging samples for complex conversations, expanding the learning space of the training model.

\subsection{Interpretation in Conversational Dense Retrieval}
\label{subsec:Interpretation_CDR}
Although conversational dense retrieval has proven effective for conversational search, a significant limitation of these approaches is low interpretability, which impedes an intuitive understanding of model behaviors and hinders targeted improvements.
To this end, the existing studies~\cite{radlinski2017theoretical} attempt to unravel the behavior characteristics of various conversational dense retrieval models.
For example, 
Mao et al.~\cite{Lecore} propose LeCoRE, which enhances interpretability by creating a denoised and sparse lexical representation of the conversational session. 
Specifically, LeCoRE extends the SPLADE model~\cite{SPLADE}, a lexical-based model to further transform latent representation into lexical representation in the vocabulary space, which provides a clear and understandable representation related to the search results.
Besides, Christmann et al.~\cite{christmann2023explainable} increase the model interpretability by combining information from multiple sources with user-comprehensible explanations for search results.
Further, Cheng et al. propose CONVINV~\cite{ConvInv} to transform the opaque, high-dimensional session embeddings generated by conversational dense retrieval models into explicit, interpretable text that can be easily understood by humans. 
The transformation is achieved by a model trained to invert the session embeddings back into a textual form that is coherent and clear and maintains the retrieval performance of the original embeddings.
Instead of explicitly conducting query reformulation for conversational search, more detailed investigations of the interpretability in dense retrieval are expected, which can avoid unfair or misleading search results. 

\subsection{Re-ranking in Conversational Search}
\label{subsec:Reranking_CS}

Re-ranking is essential after the first retrieval step in the traditional search procedure~\cite{llmir_survey} which aims to re-order the documents fetched from the first stage retrieval by enhancing the relevance of the associated query. 
Within the conversational scenarios, the same search pipeline can be adapted, i.e., combine the first-stage context representations with a second-stage re-ranker to perform the conversational search. 
Effective re-ranking is expected to significantly boost the accuracy of the subsequent answer extraction process, as it enables the reader model to work with the best possible set of documents, thus improving the overall quality and reliability of the conversational search system.

In the context of re-ranking in conversational search,
Kongyoung et al.~\cite{monoQA} design monoQA to train a model with multi-task learning simultaneously for re-ranking retrieved passages and extracting answers via machine reading to improve both overall retrieval precision and answer accuracy.
Ju et al.~\cite{ConvRerank} attempt to improve re-ranking accuracy by employing a multi-view pseudo-labeling approach that integrates diverse contextual information.
It utilizes a view-ensemble method to generate high-quality pseudo-labeled data for fine-tuning a conversational passage re-ranker. 
Hai et al.~\cite{CoSPLADE} design a re-ranking process after the conversational retrieval stage, which is enriched with keywords identified by the first-stage retriever. 
These keywords are expected to capture the helpful contextual information and re-rank the documents with useful conversational history.
Owoicho et al.~\cite{ConvSim} investigate conversational search systems by providing feedback on system responses and answering clarifying questions, based on an initial description of the user's information need.
This work discusses the potential advancements in conversational search by developing more sophisticated feedback processing modules, e.g., conversational context-originated re-ranker.

\subsection{Limitations and Discussion}

Existing studies have achieved great success in identifying relevant passages for multi-turn complex information needs.
In this section, we aim to discuss the limitations and potential solutions regarding conversational retrieval.

\noindent \textbf{Standard of Effective Pipeline.}
Though various pipelines with different mechanisms for conversational retrieval have been explored in the literature, how to establish a unified standard to construct and compare effective pipelines is still an open question, e.g., how to compare query reformulation approaches with conversational dense retrieval models. Besides, how to incorporate the multi-turn information, e.g., the historical retrieved results, into the current turn retrieval is also underexplored, as well as the necessity of deploying a re-ranker in the retrieval pipeline. 
These problems require the community to conduct further investigation on how to build a conversational retrieval pipeline with the associated historical context under a comparable standard.
Besides, the rise of LLMs further complicates the pipeline standard. Generative retrieval models~\cite{survey_genir} or end-to-end RAG systems~\cite{lewis_nips20} inherently blend retrieval with context understanding and aspects of generation, making direct comparisons with traditional query reformulation or pipelined approaches difficult. Evaluating the retrieval component in such integrated systems requires assessing not just the relevance of retrieved passages but also their utility and impact on the final generated response~\cite{salemi2024evaluating}.

\noindent \textbf{Efficiency.} Existing conversational retrieval models generally perform better than conversational query reformulation approaches, however, they usually encounter efficiency problems since the the whole long conversation is treated as input.
Such an issue would be more severe, when the backbone retrieval model is LLMs-based, i.e., with large-scale parameters.
One possible intuitive solution is employing caching mechanisms to store intermediate representations of conversation segments, which avoids reprocessing the entire conversation history for each new query.
Besides, leveraging query performance prediction techniques~\cite{DBLP:conf/sigir/MengAAR23} can help avoid redundant pipeline procedures.

\noindent \textbf{Generalizability.} 
The dense retrievers fine-tuned with conversational search data would inevitably lose the general abilities compared to the ad-hoc search retrievers and vanilla language models, either PLM-based or LLM-based. 
To enhance the generalizability of the retrievers, which can be adapted for both ad-hoc and conversational search, the researchers are encouraged to incorporate advanced instruction tuning techniques toward different tasks or develop a mixture of data for search-originated fine-tuning.

\noindent \textbf{Explainability.} 
Though some existing studies 
explore the explainability of conversational dense retrievers, the investigation of the interpretation studies is still limited.
Thorough studies on conversational retrieval models can help the community understand the behavior of both the user and the retriever, which is essential for an interactive system.
One potential solution is to conduct a comprehensive performance analysis that explores how different aspects of the conversation (e.g., topic switch and expansion utilization) affect retrieval effectiveness and user modeling.

\section{Generation in Conversational Search} 
\label{sec: generation}
The development of generative models facilitates the evolution of conversational search systems that might enrich the user experience by producing contextually aware and personalized responses~\cite{llm_survey}. As the chat mode, i.e., multi-turn conversation, becomes a common practice when leveraging LLMs for content generation, the design of the new paradigms of the conversational search system by integrating the generation module brings much more attention.
Based on powerful LLMs as generators, response generation is coupled with retrieval through Retrieval-Augmented Generation (RAG) paradigms. This integration means the generation component is not merely presenting information but actively synthesizing retrieved knowledge within the ongoing conversational context to meet the user's information needs, further blurring the lines between traditional retrieval and generation stages.

Conversational RAG~\cite{cheng2024coral} is not well-studied in the literature.
The existing studies mainly focus on single-turn RAG, which only treats a single stand-alone query or converts the multi-turn context-dependent query by reformulation technique without leveraging historical information for RAG.
In terms of the single-turn RAG paradigm, the generation component is pivotal for crafting responses that are both informed by external knowledge and coherent with the parametric ones. 
The retrieved documents usually served as external knowledge, and the generator model, usually the LLMs in existing studies, processes them with two key methodologies~\cite{RAG_survey}: context curation~\cite{zhuang_emnlp23,yang_emnlp23_prca,xu_recomp,ma_emnlp23} and generator fine-tuning~\cite{du_emnlp22,radit}, which try to adapt the consistency between IR and RAG~\cite{cuconasu2024power,wu2024easily,mo2025uniconv} and satisfy the specific requirements of the domain or task, respectively. 
Besides, fine-tuning knowledge bases or enhancing LLMs' internal knowledge could be the other alternatives to tune systems to meet specific conversational search expectations in generative paradigms. However, a single LLM, even with specific fine-tuning, could not be considered as a conversational search system without retrieval parts in this paper. This is because solely relying on a language model with parametric knowledge is not enough for information-seeking goals, and it is also the fundamental weakness of LLMs. Thus, the existing LLMs’ application requires integration with search engines (e.g., RAG), and there is no evidence showing that the functionality of the search component could be replaced.

\begin{figure}[tp]
\centering
\includegraphics[width=0.95\linewidth]{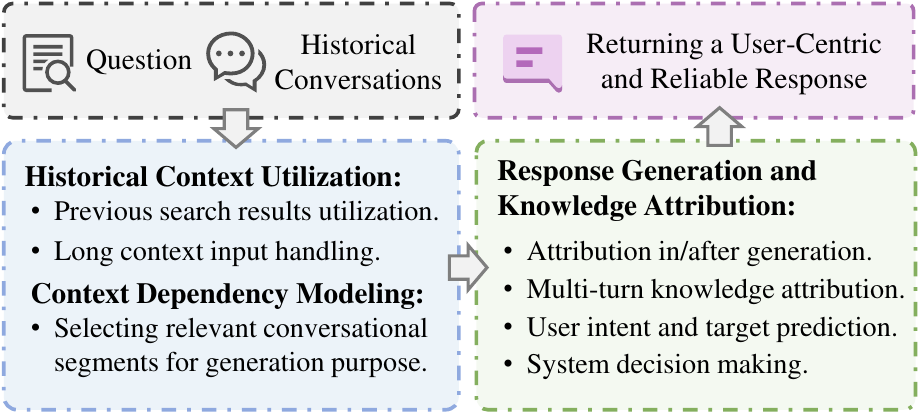}
\caption{Overview of the conversational response generation system. To achieve more user-centric and reliable responses, it is worth further exploring, including historical context utilization, context dependency modeling, and knowledge attribution.}
\label{fig:generation}
\end{figure}

Regarding the generation for conversational search, the single-turn RAG systems overlook the historical information, e.g., historical search results, the context history, turn dependency, etc.
Since conversational search systems must account for the ongoing dialogue and the evolving context of the conversation to produce responses that are not only relevant but also consistent with the user's previous interactions, historical information might be essential to generate accurate, reliable, and personalized responses. Thus, it is unclear whether the additional conversational context process can further improve the generation results, even though some mechanisms with similar goals have already been applied in the retrieval procedure.
Considering few studies are focusing on the concept of generation in conversational search, in this section, we aim to provide the potential research questions, guidance for unclear points, and corresponding solutions with different aspects to facilitate future research in this field.
An overview of the conversational generation system is illustrated in Fig.~\ref{fig:generation}.

\subsection{Utilization of Historical Search Results}
\label{sec: history search}
One of the important but overlooked historical information is the search results from previous turns, which might be helpful to the current turn RAG~\cite{ye2024boosting,Conv-CoA}. The assumption is that the retrieved results from the earlier turns could be a supplementary enhance the generation of the current turn's responses. 
The potential improvement might be attributed to increasing the recall of the current turn via improving search diversification or pushing the low-ranked documents to the top position when they frequently occur in the previous turn via enhancement. 
From a much broader perspective, the key idea is to integrate the historical search results with the current ones, by selecting the useful information and avoiding injecting additional noise.
Then, the systems are expected to construct a narrative that is not only contextually relevant but also dynamically adaptive to the evolving conversational threads, allowing the system to exhibit a semblance of episodic memory, recalling and referencing previous interactions to inform the current discourse.

Nevertheless, the principle of the usage of historical search results is unclear and more explorations are necessary in future research. Besides, two potential challenges should be taken into account when designing the approaches.
One challenge is the long context input, while the long conversational session and large top-k retrieved list will significantly increase the length of input and result in high latency costs. Besides, the context might be longer than the model input limitation. Thus, the intuitive idea is to increase the long context capacity of the RAG models and to reduce the latency by selecting or aggregating the information from the historical search results is also crucial.
Then, another challenge lies in identifying the utilization of historical search results. Since the relevance/usefulness could have some inconsistency between IR and RAG, the documents identified by the retriever might not align with the requirement of the generator. Thus, a desirable mechanism is expected to discern what information is pertinent and to synthesize it into the ongoing discourse regarding providing much more appropriate generated responses.

\subsection{Context Dependency Modeling}
\label{subsec:context_dependency_modeling}
The existing studies~\cite{mao2023LeCoRE,mo2023learning_to_relate,mo2401HAConvDR} capture context dependency in conversations in different ways to improve the accuracy of conversational retrieval by better understanding and satisfying users' search intents. The intuitive idea is to detect the query turn with similar information needs from the history.
In terms of conversational generation, the dependency between the current and historical turns might not be consistent with the impact of retrieval, which requires further exploration.

The common practice of context modeling is to simply concatenate all historical context as input, which might not be able to explicitly distinguish the relevance differences between the history and the current question. Although some sophisticated mechanisms are developed in conversational retrieval, the selection principle is not necessarily the same for the generation procedure. Thus, when conducting the context dependency for the response generation goal, the designed mechanism to avoid the interference of irrelevant information should target the final answer rather than only the ranking list quality.
Besides, existing studies for context modeling usually separate retrieval and generation as independent stages. 
However, the relevant conversational segments from the selected turns in the retrieval stage also directly affect the subsequent generation quality.
Therefore, a promising solution is to jointly learn context modeling for retrieval and generation in an end-to-end manner by utilizing the feedback of generation quality.

\subsection{Conversational Knowledge Attribution}
\label{subsec:knowledge_attribution}

Knowledge attribution, also known as citation labeling, aims to associate the output content with its knowledge sources to enhance credibility. The existing knowledge attribution methods can be divided into two categories, parallel with or after the generation~\cite{survey_LLM_attribution, survey_genir}. 
The former directly generates citations in the meanwhile with the answer generation, where the typical existing studies include WebGPT~\cite{webgpt}, LaMDA~\cite{lamda}, WebBrain~\cite{webbrain}, etc.
Aiming to generate reliable answer-evidence pairs, SearChain~\cite{Search-in-the-Chain} generates query chains to align questions and evidence, and VTG~\cite{VTG} introduces a verification module to ensure consistency between answers and evidence.
The second category~\cite{RARR,PURR,CEG} aims to first use natural language inference (NLI) models to calculate relevance between response and evidence after the generation and then post-edit the content to add citations. 

Although these methods enhance knowledge attribution and interpretability from different perspectives, they are not designed for conversational scenarios.
Thus, it is unclear whether they can be applied to the conversational generation. For example, could these methods also be attributed to the historical evidence retrieved when injecting context information from previous turns? The potential exploration and improvement could follow the aspects below.

\noindent \textbf{Multi-turn Knowledge Attribution.}  
Existing methods mainly focus on single-turn QA, while multi-turn dialogues should also consider the context dependencies and progressive knowledge attribution from different historical turns. Graph-based or hierarchical structured representations can be explored to capture topic flow and knowledge evolution. The key point is to identify the useful knowledge from the conversational history, including but not limited to historical search results, turn dependency, etc. The determined useful knowledge can be served as evidence without noise for better quality of response generation.

\noindent \textbf{User Intent and Target.} 
Knowledge attribution in multi-turn scenarios can be combined with dialogue intents and targets.
Since the user intents and target might change dynamically during the conversation, dynamic adjustment of attribution granularity and scope based on intent and target shift are required.
The potential solution is to identify the intent transition, detect the key knowledge within different user intents, and achieve intent-driven attribution.  

\noindent \textbf{System Decision Making.}
Multi-turn knowledge attribution can collaborate with the decision making of the system as the conversation going. For example, it helps to dynamically decide whether to follow up, correct errors, eliminate noise, or introduce new topics based on the certainty and importance of the attributed knowledge, which can also contribute to building up a mixed-initiative conversational search system. The knowledge attribution could serve as the foundation for different decision making strategies, while these strategies could also guide the knowledge attribution method and the granularity of the knowledge. The collaborative optimization is expected to improve both the knowledge attribution component and the intelligence of decision making of the system.

\section{Domain-specific and User-centric Conversational Search}
\label{sec:domain}

\begin{figure*}[]
  \centering
  \includegraphics[width=.9\linewidth]{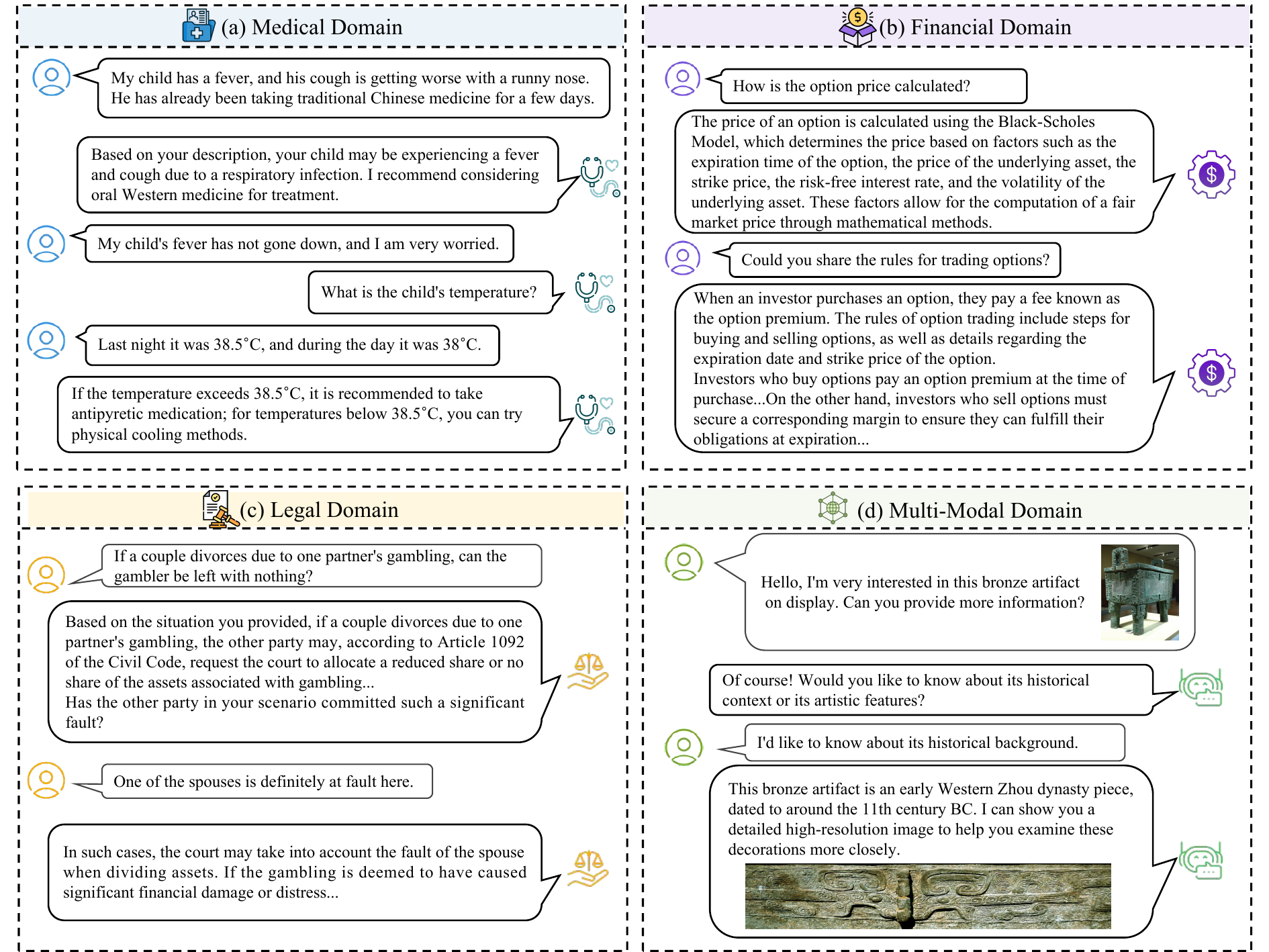}
  \caption{Typical examples of conversational search systems across four different domains: (a) Medical, (b) Financial, (c) Legal, and (d) Multi-modal. Each of them requires specific information processing.}

  \label{fig:multi-domain}
\end{figure*}

The rapid development of conversational search systems facilitates various applications in the general and diverse specific domains, including medical~\cite{tian2024opportunities}, finance~\cite{lee2024survey}, law~\cite{liu2023investigating}, e-commercial~\cite{bernard2023mg}, etc.  
The domain-specific systems might store the corresponding knowledge in different forms, e.g., tables, graphs, or multi-modal, which require the search systems to gather various information resources via reasoning skills.
Aside from specific domain requirements, the system should cover the intended utilization of the conversational search model by real users.
Thus, the designed system should also consider a user-centric perspective~\cite{wang2024user} in both dataset construction and evaluation criteria rather than only focusing on predefined search abilities.
Besides, from the personalized perspective, since the backgrounds of users are different, the search results from the conversational systems should yield various results for the same query according to the user profiles with identified relevant information for each query turn~\cite{aliannejadi2024trec}.
In this section, we present the pioneering conversational search studies conducted in various specific domains and user-centric scenarios, and we conclude by offering a comprehensive discussion of the emerging limitations and challenges.

\subsection{Domain-specific Conversational Search}
\label{subsec:domain-specific}
Earlier studies~\cite{le2020uniconv,jeon2021domain,zhao2023multi} attempt to solve domain-specific issues via multi-domain/task learning based on task-originated conversation system.
The key idea is to search cross-domain information for sharing by reasoning over the conversational history.
Though these systems can somehow process various types of information across domains, they cannot achieve satisfactory performance when targeting each specific domain. Thus, various domain-specific models are necessarily proposed for each domain.

\subsubsection{Medical Domain.}
\label{subsubsec:Medical_Domain}
The common practice of establishing a conversational search system in the medical domain is to extract relevant medical concepts from retrieved information within a conversational interface for response generation~\cite{xia2022medconqa}, which is illustrated in Fig.~\ref{fig:multi-domain} (a). 
The medical domain presents unique challenges due to the significant consequences at stake, the need for extreme accuracy, the frequent use of complex terminology that is often unclear to non-experts, and privacy concerns. 
Nevertheless, existing conversational search is well-suited here because medical conditions often require discussing history, symptoms, and treatments over time, fitting the conversational paradigm.
The variability in how medical concepts are expressed (i.e., the same concept might be written in numerous non-canonical forms) poses challenges for accurately searching texts with relevant concepts and for modeling conversational context, particularly with co-reference and omission~\cite{bhowmik2021fast}.
Thus, effective systems must bridge the gap between patient language and clinical terminology.

\noindent \textbf{LLM-based approaches.}
With the emergence of LLMs, the medical domain conversational search systems paradigm encounters new opportunities and challenges~\cite{zhou2023survey,tian2024opportunities}. 
The available medical corpus~\cite{han2023medalpaca,labrak2024biomistral} enables continual training of the existing LLMs to enhance their superb capability in the associated domain.
For example, Toma et al.~\cite{toma2023clinical} propose an open-source expert-level medical language model by encoding the knowledge based on conversational context. Bao et al.~\cite{bao2023disc} construct high-quality supervised fine-tuning datasets based on dialogue and knowledge graphs to bridge LLMs with real-world medical consultation.
However, the generative LLMs remain an unyielding concern in returning accurate responses, especially when the stored knowledge is out-to-date.
To mitigate such issues, Pal et al.~\cite{pal2023med} propose medical domain hallucination detection approaches for LLMs to reduce inaccurate responses and Wang et al.~\cite{wang2024jmlr} integrate the search components into medical LLMs to help retrieve external knowledge for updating support evidence, which demonstrates domain retrieval is more important than domain pre-train.
Although integrating external knowledge is promising, the current LLM-based approaches still struggle with hallucination issues, which is especially dangerous in medicine. Furthermore, existing medical domain benchmarks often focus on standardized question-answering tasks rather than simulating realistic, nuanced patient consultations. Evaluating safety, reliability, and the ability to handle sensitive information appropriately remains critical and underdeveloped~\cite{degachi2025towards}.
Thus, there is a significant lack of effective methodologies to enhance the LLMs to achieve users' specific goals in medical settings.

\noindent \textbf{Language-specific knowledge in Medical Domain.}
Considering the language-specific knowledge that is highly relevant to local cultural, historical, political, and regional contexts, another crucial factor for the medical domain conversational search system is conducting multi/cross-lingual information retrieval~\cite{nie2022cross,huang2024survey}. For example, with its rich history deeply connected to cultural heritage, forms a unique medical search system~\cite{cai2024medbench,yang2024zhongjing}. 
While a translation system can address some of these issues~\cite{jin2024better}, it may prove ineffective due to the complexity of medical terminology and the challenges in achieving precise translations. 
Integrating diverse medical knowledge, e.g., for Traditional Chinese Medicine, requires more than translation, which demands culturally aware models capable of understanding different diagnostic paradigms.
Thus, exploring the fundamental mechanism of integrating language-specific medical knowledge and conversational search is promising and unexplored~\cite{alonso2024medexpqa}.

\subsubsection{Financial Domain.}
\label{subsubsec:Financial_Domain}
Finance with numerical data demands high precision and often involves complex reasoning over structured and unstructured information, such as tables and reports~\cite{liu2021finbert,huang2023finbert}. The information needs of the users might be real-time data or performing transactional tasks. Conversational search systems can be suitable for guiding users through complex financial products or explaining market trends, but must handle numerical reasoning robustly and ensure information timeliness~\cite{chen2022convfinqa}.
A series of studies have been conducted in the financial domain in the context of conversational search. To simulate the real-world scenario, Sharma et al.~\cite{sharma2021stockbabble} build up a demo to show how financial agents can support stock market investors under conversational scenarios as shown in Fig.~\ref{fig:multi-domain} (b).
Aim to develop sophisticated information systems and evaluate the model automatically, the follow-up studies focus on constructing the datasets based on conversational question answering tasks and storing the financial information in tabular~\cite{deng2022pacific,liu2023tab}. The key challenge lies in retrieving the relevant information from the structured data and generating the final answer by multi-hop reasoning, however, the performance of the models is still unsatisfactory because of data scarcity~\cite{chen2022convfinqa}.

With a huge pre-trained corpus, LLMs show superior capacity in reasoning ability but still fail in the financial domain, especially requiring retrieving numerical information~\cite{chen2022convfinqa}.
To address such issues,~\cite{choi2023conversational} develop an LLM-based conversational search model for the financial domain, tailored for query intent classification and knowledge base labeling. However, the system has not been subjected to comprehensive information retrieval testing due to a lack of precise instructions and evaluation protocol. Xie et al.~\cite{xie2024pixiu} further propose a comprehensive benchmark, instruction dataset, and LLMs for finance, facilitating the research field by providing systematic resources. 
Overall, current systems show promise but often struggle with multi-step numerical reasoning and integrating real-time data feeds. The existing datasets, benchmark, and evaluation protocol are valuable but may not fully capture the complexity and dynamism of real-world financial queries. These factors constrain the development of financial systems to satisfy users' special information needs~\cite{lee2024survey}. Moreover, ensuring that the generated advice is both accurate and compliant with financial regulations represents a significant challenge that has been largely neglected in existing research evaluations.


\subsubsection{Legal Domain.}
\label{subsubsec:Legal_Domain}
Legal case retrieval is rapidly gaining prominence in the IR community due to its specific social impact, which aims to search for supporting evidence for a given query case. However, using a legal search system is challenging for those lacking domain knowledge because the user cannot fully express their specific information needs as illustrated in Fig.~\ref{fig:multi-domain} (c). 
Specifically, the legal domain requires retrieving precise supporting evidence, e.g., statutes, case law, which often involves complex logical reasoning and the interpretation of highly technical, terminology-rich texts. 
The ambiguity in law cases is common, while interpretations have significant impacts.
To this end, Liu et al.~\cite{liu2021conversational} try to deploy a conversational search system into legal case retrieval and demonstrate its better effectiveness than the traditional ones. This is also attributed to the domain-specific linguistic phenomena, where the systematic patterns in language use, as explained by certain theories of conversational implicature, can sometimes explain why judges interpret legal texts in ways that differ from their literal meanings~\cite{slocum2016conversational}. To further improve the performance of the system, a series of studies address different aspects, including developing agent-based action decision-making mechanisms~\cite{liu2022query,liu2023investigating}, re-rank long-context response~\cite{askari2023closer}, and learn to ask clarification questions as introduced in Section~\ref{sec: SC-legal}. 
While achieving some improvements, current conversational search in the legal domain faces challenges in ensuring the retrieved evidence is contextually appropriate and legally sound. The ability to handle long, complex legal documents effectively within a conversational context~\cite{askari2023closer} is also a key area for improvement.

\subsubsection{Other Domains.}
\label{subsubsec:Other_Domains}
In addition to the aforementioned domains with text data only, the conversational search systems also facilitate domains with different data forms. For example, the e-commercial domain involves various item attribution and user profile information, which requires dealing with product search, preference recommendation, and question answering based on item catalog matching, user modeling, and item-related knowledge understanding, respectively. 
Some recent studies~\cite{jia2022convrec,bernard2023mg,zamiri2024benchmark} contribute to resource construction for the mixture search goals, entity retrieval, and scaling a broad-coverage dialogue corpus, which enables further exploration to become much more feasible.

Besides, aiming to retrieve from multiple resources, Liao et al.~\cite{liao2021mmconv} extend the conversational session to multi-modal scenarios, requiring the text and the image information for solving the downstream tasks. Li et al~\cite{li2022mmcoqa} emphasize more on the search task and further incorporate tabular evidence as retrieval sources. A typical example is shown in Fig.~\ref{fig:multi-domain} (d). The challenging task raises the difficulty of the alignment and reasoning among the retrieved evidence from different modalities, resulting in unsatisfactory performance by the previous state-of-the-art methods. To address it, MOQAGPT~\cite{zhang2023moqagpt} is proposed based on a divide-and-conquer strategy to integrate the ranking results from various resources.
More studies are required to facilitate this specific task setting.

\subsection{User-centric Conversational Search}
\label{subsec:user-centric}
Conversational search systems are effectively employed across various scenarios to support users in seeking information. Instead of mainly focusing on specific predefined model abilities, such as retrieving world knowledge, reasoning across various candidate evidence, etc, some existing studies~\cite{chu2022convsearch} concentrate on the perspective of user satisfaction, i.e., to improve users' satisfaction by addressing their information needs under distinct intents.
This is non-trivial because explicitly requesting satisfaction feedback from users is challenging and the approaches employed in traditional IR have not been thoroughly investigated in the context of conversational search.
To facilitate the studies on user-centric conversational search, some benchmarks~\cite{chu2022convsearch,wang2024user} have been proposed recently for evaluating the system from a user-centric perspective and investigate how to build up proactive conversational search systems~\cite{erbacher2024paqa,meng2025bridging} together with human-centered agents~\cite{deng2024towards}. Besides, some existing studies~\cite{yan2023titan,owoicho2022trec} focus on mixed-initiative interactions, aiming to provide a better experience with multiple feedback to the user. Thus, maintaining a user-centric system is essential to ensure these techniques continue to serve users' interests in future development.

\begin{table*}[!t]
\centering
\scalebox{0.9}{
\begin{tabular}{lccccccc}
\toprule
\multirow{2}{*}{Dataset} &  Release &  \multirow{2}{*}{Domain} & \multirow{2}{*}{Task} & \multirow{2}{*}{Language} & \multirow{2}{*}{Data Form} & \multicolumn{2}{c}{Size} \\
\cmidrule(r){7-8}
& Time & & & & & Train & Test\\
\midrule
MedConQA~\cite{xia2022medconqa} & 2022 & Medical & QA & zh & Text, KG & 1.3M & -\\
MedAlpaca~\cite{han2023medalpaca} & 2023 & Medical & QA & en & Text & 67k$^{\ddagger}$ & 374$^{\ddagger}$\\
Disc-medllm~\cite{bao2023disc} & 2023 & Medical & QA & en & Text, KG & 400k$^{\ddagger}$ & 1.6k$^{\ddagger}$\\
BioMistral~\cite{labrak2024biomistral} & 2024 & Medical & QA & en & Text & 367k$^{\ddagger}$ & 7k\\
Medbench~\cite{cai2024medbench} & 2024 & Medical & QA & en, zh & Text & - & 39k$^{\ddagger}$\\
MedExpQA~\cite{alonso2024medexpqa} & 2024 & Medical & QA & en, es, it, fr  & Text & 434 & 125\\
CinvFinQA~\cite{chen2022convfinqa} & 2022 & Finance & QA & en & (Numerical) Text & 3k & 434 \\
Tab-CQA~\cite{liu2023tab} & 2023 & Finance & QA & zh & Text, Table & 101k & 3.6k \\
ConvLCR~\cite{liu2021conversational} & 2021 & Legal & Case Search & zh & Text & - & 14 \\
LexRAG~\cite{li2025lexrag} & 2025 & Legal & QA & zh & Text & - & 5k \\
E-ConvRec~\cite{jia2022convrec} & 2022 & E-commerce & Product Search \& Rec. & zh & Text, User, Product & 620k & 7k \\
MG-ShopDial~\cite{bernard2023mg} & 2023 & E-commerce & Product search \& Rec. \& QA & en & Text, Product & - & 64 \\
MMconv~\cite{liao2021mmconv} & 2021 & Multi-modal & Search & en & Text, Image & 31.8k & 3.9k\\
MMcoqa~\cite{li2022mmcoqa} & 2022 & Multi-modal & Search \& QA & en & Text, Image, Table & 4.6k & 575 \\
Qilin~\cite{chen2025qilin} & 2025 & Multi-modal & Search & zh & Text, Image, Product & - & 5.7k \\
ConvER~\cite{zamiri2024benchmark} & 2024 & Multi-modal & Entity Retrieval & en & Text, KG & 14k & 1.6k\\
ConvS~\cite{chu2022convsearch} & 2022 & User-Centric & User Behaviour & zh & Text, User, Click & - & 8k\\
TREC iKAT 2023~\cite{aliannejadi2024trec} & 2023 & User-Centric & Personalized Search & en & Text, User & - & 176\\
TREC iKAT 2024~\cite{abbasiantaeb2025conversational} & 2024 & User-Centric & Personalized Search & en & Text, User & - & 103\\
LAPS~\cite{joko2024doing} & 2024 & User-Centric & Personalized Rec. & en & Text, User & 19k & -\\
URS~\cite{wang2024user} & 2024 & User-Centric & User Satisfactory & en, zh & Text & - & 1.8k\\ 
\bottomrule
\end{tabular}}
\caption{An overview and statistical details of the domain-specific conversational search data resource. $\ddagger$ denotes several subsets associated with different sub-tasks are contained.}
\label{table: multi-domain resources}
\end{table*}

Another research line attempts to facilitate user-centric conversational systems via personalized search, which requires satisfying users' information needs and considering their backgrounds when searching candidate documents~\cite{aliannejadi2024trec,abbasiantaeb2025conversational}. Specifically, the search results from the systems should yield different results for the same query according to the user profiles.
The primary challenge is the issue of data scarcity in model development. To this end, Joko et al.~\cite{joko2024doing} construct a large-scale personalized multi-session conversational search dataset synthetically by LLMs. Mo et al.~\cite{mo2024leverage} explore how to integrate query rewriting with personalized elements in a zero-shot manner by LLMs. Some recent studies aim to define concrete scenarios and evaluate the personalized ability of the systems~\cite{deng2024multi,aliannejadi2024trec}. With the development of the data resource, more sophisticated personalized systems are expected.

\subsubsection{Ethical Considerations and Bias}
A user-centric conversational search system should take the aspect of ethical considerations and potential bias into account.
Since the LLMs are trained on vast web data, LLM-based conversational systems could inherit and propagate societal biases related to gender, race, socioeconomic status, and other attributes~\cite{lajewska2024can}. As a result, the skewed or unfair information presentation might be presented in search results or generated responses.
This bias would be amplified, especially when it is connected with user profiles or interaction logs~\cite{chen2023customized}.
Besides, conversational interactions can involve users revealing sensitive personal information.
As conversational search systems will be widely deployed in the future and used by daily users more frequently, systems must employ robust privacy-preserving techniques~\cite{lin2024federated} for data handling, storage, and model training, especially when personalization is involved.

\subsection{Data Resources}
The main bottleneck for developing domain-specific conversational search systems lies in data scarcity, which requires the knowledge resources to be injected into the models or serve as the external base. Besides, the available benchmark is crucial to evaluate the system's specific ability corresponding to the requirements of various domains. To the best of our knowledge, we present the statistical detail of the data resources from each domain in Table~\ref{table: multi-domain resources}, which reflect the development of each special domain and provide the feasibility to train and evaluate the more sophisticated models. 

\subsection{Limitations and Discussion}
Despite the remarkable efforts on domain-specific conversational search, further exploration is necessary. We provide a brief discussion of the current limitations and potential improvement as follows.

\noindent \textbf{Data Curation.}
The challenge in domain-specific tasks mainly lies in the lack of a comprehensive document repository, which constrains the feasibility for model training and evaluation. Compared with the general domain, domain-specific data are much higher cost to obtain due to the requirement of domain knowledge for the annotator. Thus, large-scale resources are desirable for domain-specific development and the automatically data generation techniques, e.g., LLM-aided, could be a potential avenue~\cite{abbasiantaeb2025improving,meng2025query}.

\noindent \textbf{Domain-Specific IR.}
The complex terminology used in specific domains increases the difficulty of matching the retriever and query semantics, especially when involved with local cultural, historical, political, and regional backgrounds. This could become extremely severe when the user lacks of expert knowledge and expresses the terminology in different ways to query. One potential solution is to leverage the domain-specific LLMs to help reformulate queries. Besides, the further development of domain-originated search engines by considering multi-factors are important.

\noindent \textbf{Faithfulness.}
The reliable results are important in domain-specific applications. For example, disseminating incorrect or deceptive information can lead to greater ethical and practical implications than in other sectors.
The misinformation, disinformation, and hallucination in retrieval and generation increase the difficulty for the search systems to return faithful answers, which requires more investigation.

\section{Benchmark and Evaluation}
\label{sec:evaluation}
\begin{table*}[!t]
\centering
\scalebox{0.9}{
\begin{tabular}{lcclcccc}
\toprule
Name & Release Time & Language & Task & RAG & TS & Mixed-initiative \\
\midrule

MSDialog~\cite{qu2018analyzing} & 2018 & en & User Intent, Search & \xmark & \xmark & \xmark \\

TREC CAsT-19~\cite{dalton2020cast} & 2019 & en & QR, Search & \xmark & \xmark & \xmark\\
Qulac~\cite{aliannejadi2019asking} & 2019 & en & SC & \xmark & \xmark & \cmark \\
CANARD~\cite{elgohary2019can} & 2019 & en & QR & \xmark & \xmark & \xmark \\
MANtIS~\cite{penha_mantis} & 2019 & en & Search, Generative QA & \cmark & \xmark & \xmark \\
WoW~\cite{dinan_iclr19_wizard} & 2019 & en & Search, Generative QA & \cmark & $\bigtriangleup$ & \xmark \\
CLAQUA~\cite{xu2019asking} & 2019 & en & Clarification for QA & \xmark & \xmark & \cmark \\
TREC CAsT-20~\cite{dalton2021cast} & 2020 & en & QR, Search & \xmark & \xmark & \xmark\\
OR-QuAC~\cite{qu2020open} & 2020 & en & Search, Extractive QA & \cmark & \xmark & \xmark \\
MIMICS~\cite{zamani2020mimics} & 2020 & en & Web Search Clarification & \cmark & \xmark & \xmark \\
ClarQ~\cite{kumar2020clarq} & 2020 & en & Clarification for QA & \xmark & \xmark & \cmark \\
ClariQ~\cite{aliannejadi2020convai3} & 2020 & en & SC & \xmark & \xmark & \cmark \\
ClariQ-FKw~\cite{aliannejadi2020convai3} & 2021 & en & SC & \xmark & \xmark & \cmark \\
WISE~\cite{ren2021wizard} & 2021 & en & User Intent, SC, Search, Generative QA & \cmark & \cmark & \cmark\\
QReCC~\cite{anantha2021open} & 2021 & en & QR, Search, Generative QA & \cmark & \xmark & \xmark \\
TopiOCQA~\cite{adlakha2022topiocqa} & 2021 & en & Search, Generative QA & \cmark & \cmark & \xmark\\
TREC CAsT-21~\cite{dalton2022cast} & 2021 & en & QR, Search & \xmark & $\bigtriangleup$ & \xmark\\
MultiDoc2Dial~\cite{feng_emnlp21_multidoc2dial} & 2021 & en & Search, Generative QA & \cmark & \cmark & \xmark\\
Abg-CoQA~\cite{guo2021abg} & 2021 & en & Clarification for QA & \xmark & \xmark & \cmark \\
SaaC~\cite{ren_serp_saac_case} & 2021 & en & Search, Generative QA & \cmark & \xmark & \xmark\\
ConvS~\cite{chu2022convsearch} & 2022 & zh & User Behaviour, Search & \xmark & \xmark & \xmark\\
TREC CAsT-22~\cite{owoicho2022trec} & 2022 & en & QR, SC, Search, Generative QA & \cmark & $\bigtriangleup$ & \cmark\\
MIMICS-Duo~\cite{tavakoli2022mimics} & 2022 & en & Web Search Clarification & \cmark & \xmark & \xmark \\
Doc2Bot~\cite{fu_emnlp22_doc2bot} & 2022 & en & Search, Generative QA & \cmark & \xmark & \xmark\\
FOCUSQA~\cite{Barlacchi_emnlp22_focusqa} & 2022 & en & Search & \xmark & \cmark & \xmark\\
TREC iKAT-23~\cite{aliannejadi2024trec} & 2023 & en & Personal., QR, Search, Generative QA & \cmark & \xmark & \cmark\\
ConvRAG~\cite{ye2024boosting} & 2023 & zh & QR, Search, Generative QA & \cmark & \cmark & \xmark\\
StatCan Dialogue~\cite{lu_eacl23_StatCan_Dialogue_Dataset} & 2023 & en,fr & Search, Generative QA & \cmark & \xmark & \xmark\\
DomainRAG~\cite{wang2024richrag} & 2024 & zh & Search, Generative QA & \cmark & $\bigtriangleup$ & \xmark\\
ProCIS~\cite{samarinas2024procis} & 2024 & en & Search & \xmark & \cmark & \cmark\\
Melon~\cite{yuan2024asking} & 2024 & en & Multi-modal Clarification & \xmark & \xmark & \cmark \\
Coral~\cite{cheng2024coral} & 2024 & en & Search, Generative QA & \cmark & \cmark & \xmark \\
\bottomrule
\end{tabular}}
\caption{An overview of conversational search benchmark with various evaluation settings. \textbf{QR}, \textbf{SC}, \textbf{QA}, and \textbf{TS} denote query reformulation, search clarification, question answering, and topic-switch, respectively.}
\label{table: benchmark}
\end{table*}

A complete and intelligent conversational search system should be integrated with multiple components, including query reformulation, search clarification, conversational retrieval and response generation, etc. With the flexible interactive interface of the conversational search system, the information return form could depend on specific queries, which could be a rank list of the relevant documents, a summarization of them, a retrieval-augmented generated response, etc.
The evaluation of existing conversational search systems can be divided into two categories~\cite{chu2022convsearch}: 1) based on ranking metrics (e.g., MRR, NDCG, Recall), or 2) based on the semantic similarity metrics (e.g., F1, BLUE, and ROUGE), where the former is usually applied to retrieval tasks while the latter is commonly used with generation tasks.
In this section, we first present different aspects of evaluation in existing studies with summarized available benchmarks. 
Then, we discuss the overlooked issues in retrieval-based and generation-based evaluation and propose some potential directions for improvement.

\subsection{Evaluation Overview}
\label{subsec:evaluation-overview}
The overview of the existing conversational search benchmarks is presented in Table~\ref{table: benchmark},
where each benchmark corresponds to a specific setting, e.g., whether requiring to generate a final answer (RAG), involving a topic-switch (TS) phenomenon, or multiple reaction goal (Mixed-initiative) in a conversation.
We can find that in the era of the large language model (after 2021), the user-iterative scenario and the RAG task have become more important and popular. Thus, different from the single-turn setting that most focuses on the accuracy of the results, it is crucial to evaluate the retrieval and generation task with various aspects from a conversational perspective.
Besides, the widespread exposure to powerful LLMs has significantly shifted user expectations and behaviors, e.g., from one-shot search to multi-turn interactions, which raises the new requirement for developing evaluation protocols from the aspect of users.

\subsubsection{Retrieval-based Evaluation}
\label{subsubsec:Retrieval_based_Evaluation}
Following the traditional single-turn ad-hoc retrieval, existing studies on conversational retrieval commonly treat every query turn within a conversational session as a single query with the corresponding relevance judgment as ground truth. 
This is because the available conversational search datasets are constructed by inheriting from the single-turn datasets. 
In addition to the traditional ad-hoc search evaluation and focus on conversational features, Fu et al.~\cite{fu2022evaluating} examine whether the same pattern of the assessment in ad-hoc search exists in conversational search based on both the user satisfaction and the observed performance of the search. 
Li et al.~\cite{li2022ditch} and Siblini et al.~\cite{siblini2021towards} argue that the existing benchmarks using ground-truth answers provided in historical context are unreliable to evaluate the developed models regarding real human-machine conversations.
To simulate an open-ended environment with unpredictable conversation trajectories for the practical situation, a series of studies leverage user simulators to provide feedback
to the system’s responses.
Early works~\cite{salle2021studying,lipani2021doing} study how to estimate the effectiveness of conversational search based on user simulation. Then, Wang et al.~\cite{wang2022simulating} simulate the risk of clarifying user questions and develop a risk-aware control model in terms of the search experience.
To further facilitate the simulation of user feedback for conversational search, some recent studies~\cite{wang2023depth,owoicho2023exploiting,balog2023user,bernard2024leveraging} develop user simulator-based frameworks to build up more user-system interaction as mixed-initiative to improve the conversational search systems via specific user feedback. Besides, it can also be used to evaluate mixed-initiative conversational search systems~\cite{sekulic2022evaluating}. 
The goal of mixed-initiative systems is to enrich the interaction pattern, which can take initiative at any moment in the conversation and offer the user various questions or suggestions to address the information needs eventually.
To evaluate the mixed-initiative aspect, Aliannejadi et al.~\cite{aliannejadi2021analysing} analyze the potential for more engaging conversational searches and the difficulty of assessing them due to the broader interaction scope. To this end, Sekulic et al.~\cite{sekulic2022evaluating} achieve automatic evaluation by proposing a conversational user simulator. Besides, some recent studies~\cite{meng2023system,fu2023priming} focus on analyzing the prediction of system initiative and user action, which provide a new perspective for evaluating mixed-initiative systems.

\subsubsection{Generation-based Evaluation}
\label{subsubsec:Generation_based_Evaluation}
The flexible user-system interaction provides the chance to address complex information needs by multi-turn querying, which means the requirement of the user might not only be a rank list but also a concise answer from the retrieved results. 
Such a paradigm has become significantly more popular in the era of LLMs, making generation-based evaluation essential. 
However, conventional metrics such as F1, BLEU score, and exact match (EM) might not fully capture the quality of the generation. On one hand, some reasonable and correctly generated answers might not be annotated as the evaluation gold standard and fail in the traditional token-level metrics. On the other hand, the LLMs can 
yield accurate responses despite imperfect retrievers or hallucinations with semantic relevant retrieved text~\cite{wu2024easily}, raising the issue of evaluating the system by taking into account both retrieval and generation.
Aim to combine the top retrieved documents into a comprehensive, pertinent, and concise response, Lajewska et al.~\cite{lajewska2023towards} collect snippet-level answer annotations for existing conversational search benchmarks to fill the evaluation gap between retrieval and generation.
To improve the correlation of retrieval effectiveness with the performance of the generation procedure, Salemi et al.~\cite{salemi2024evaluating} propose the eRAG metric, which evaluates the utilization of each retrieved document based on the ground truth labels of downstream tasks.
Nevertheless, the evaluation gap remains between the aspect of IR and RAG, i.e., an IR-relevant document might not be effective for the downstream RAG task~\cite{cuconasu2024power}. Thus, how to assess the performance of retrievers within LLM-based generators and better align the retriever-generator with the overall conversational search system is still an open challenge, which requires further exploration~\cite{alinejad2024evaluating}.

\subsubsection{User-based Evaluation}
\label{subsubsec:User_based_Evaluation}
Except for delving into specific metrics for accuracy, it is necessary to consider the practical impact, such as user satisfaction between traditional search interfaces and conversational search interfaces. 
The property of mixed-initiative of conversational search systems and their interfaces provides potential for higher user engagement and satisfaction, particularly for complex, exploratory information needs, which is inapplicable in traditional one-shot search.
The multi-turn interactions can help refine queries and synthesize information effectively, which provides a different search experience.
Thus, evaluating conversational systems with careful consideration of user experience and satisfaction within realistic interaction contexts is important. 
The previous studies in human-computer interaction~\cite{preece1994human,degachi2025towards} and interactive information retrieval~\cite{aliannejadi2024interactions} aim to investigate how to maximize user utility~\cite{trippas2017people} by emphasizing more on qualitative user studies, understanding cognitive load, interaction friction, and the nuances of dialogue success beyond simple definitions of relevance in search. 
In the era of LLMs now, users often expect more direct and synthesized answers, engage in longer and more exploratory dialogues, and may deploy the conversational intelligent systems for tasks beyond simple fact retrieval, such as task completion or content creation. 
Thus, the users might have higher expectations regarding the naturalness and coherence of the conversation, which consequently results in the requirement to evaluate user satisfaction or conduct user studies.
Some recent studies aim to develop user-based evaluation standards from the aspects of user satisfaction in the task completion~\cite{wang2024user,meng2024query}, the interactions in the command-and-control paradigm~\cite{trippas2024re}, and detect-then-mitigate cognitive bias~\cite{ji2024towards}.

\subsection{Limitations and Discussion}
Although existing conversational search models have achieved significant development, the convincing evaluation of these systems is still an open question. The main challenge lies in how to evaluate them with several aspects 
as follows.

\noindent \textbf{Conversational-Originated Evaluation.}
The existing conversational search systems are developed and evaluated based on the traditional search metric for each query turn independently. Besides, the current benchmarks assume the previous turns are answered correctly to ensure conversation consistency~\cite{li2022ditch}, which is not practical for real-world applications. Samarinas and Zamani~\cite{samarinas2024procis} first attempt to propose the npDCG metric as well as a new benchmark to evaluate a system from both reactive and proactive aspects. The exploration of user simulation (with LLMs) tries to make the conversational system evaluation much more realistic and user-centric. More evaluation metrics and simulators are expected to evaluate the system at the conversational level and can reflect user experience.

\noindent \textbf{Retrieval Effectiveness Evaluation.}
The development of generative models requires the system to craft a final answer, e.g., RAG, in some cases based on the search results. However, the demands of retrieval and generation might not align, the retrieval aims to address the relevance of the retrieved set while the generation needs to maximize the utility of retrieved results for downstream tasks. Thus, the new evaluation protocols toward bridging the usefulness between retrieved results and generation goals are expected. The potential solution is to develop some determination mechanisms with suitable objectives to identify the useful pieces for the generator on top of the retrieved results.

\noindent \textbf{Mixed-initiative Evaluation.}
The conversational interface enables the flexibility of the return form, which could be a clarification question, a list of ranked documents, a direct/retrieval-augmented precise response, etc. 
Thus, an intelligent conversational search system should be able to determine the suitable decisions and forms of information generation return to the user, which is consistent with mixed-initiative evaluation in the literature~\cite{owoicho2022trec}.
Though the LLM agents~\cite{xi2023rise,gong2024cosearchagent} have the basic ability to plan the decision-making, a more precise mechanism is desirable. The main challenge lies lack of benchmarks to evaluate the effectiveness of each turn action toward final information needs in an end-to-end manner.
A possible solution is first to develop a benchmark and propose the evaluation methods for the accuracy of action prediction and the contribution for information seeking.

\noindent \textbf{User Satisfaction Evaluation in Conversation and Interaction.}
The general evaluation metrics focused solely on document relevance or turn-level answer accuracy are insufficient for modern conversational search systems.
The evaluation toward holistic, task-based, user-centric metrics in the context of LLM-based systems could be connected with the findings from decades of human-computer interaction and interactive information retrieval research~\cite{trippas2025applying}, especially user behaviors and expectations could be involved~\cite{trippas2024adapting}.
Besides, compared to traditional ad-hoc search, users might notice conversational systems are slower for simple fact retrieval, producing verbose responses, or have difficulties adapting to a multiple-turn-based interaction model compared to the familiar keyword-and-scan approach of traditional search. 
The higher latency, trustworthiness in LLM-generated answers, and the system's ability to gracefully handle errors significantly impact user satisfaction and adoption rates.
Thus, developing more meaningful evaluation protocols for current powerful, but complex, LLM-based conversational search systems is promising and desirable.
\section{Conclusion and Future Directions}
\label{sec:conclu}
In this survey, we have conducted an in-depth review of conversational search systems, focusing on four essential modules: query reformulation, search clarification, conversational retrieval, and response generation. For each component, we reviewed the current methodologies and advancements and discussed potential avenues for future developments. A significant portion of our discussion has centered on the impact of LLMs on each critical module of conversational search systems. We explored how LLMs can be integrated and support these components, offering new opportunities with emerging challenges for innovation and performance improvement.
Moreover, we discussed the application of existing conversational search systems in user-centric, domain-specific scenarios, providing insights into their practical use cases. We also summarized the available benchmark resources and evaluation methods for assessing the performance of conversational search systems and indicated the potential improvement toward more practical evaluation techniques.

We expect our effort in this survey can illustrate the required landscape of modern conversational search and provide a strong foundation for future research in this rapidly evolving field.
Aiming for more intelligent conversational search systems for future development, the overall suggestions and research questions 
are summarized as follows:

\noindent \textbf{Intelligent Decision Making.} 
How to enable conversational search systems to robustly determine the optimal action, e.g., clarify, retrieve, generate, invoke an external tool, at each conversational turn, adapting dynamically based on dialogue state, user intent, and system confidence?
The appropriate system returns as the interactions with the user are the guarantee for a good search experience, along with the conversation diving in.
An alternative is to leverage LLM agents to complete complex decision tasks with tool-call, advanced memory, reasoning, and planning capabilities. The challenges lie in ensuring relevance and accuracy, mitigating biases, providing real-time responses, etc.

\noindent \textbf{Faithful and Abundant Resources.} 
How to develop reliable knowledge attribution methods for multi-source conversational generation to mitigate hallucination?
Faithfulness is crucial for search systems, which is an open challenge for generative-based systems. The key issue is to ensure the generated content are reliable response for the user, by indicating resources for knowledge attribution. Besides, to make resources abundant, e.g., integrating multi-modality information, can also increase credibility and improve the results diversification of the interactive systems.

\noindent \textbf{Proactive and Personalized Conversation.} 
How can we enable the conversational search systems to proactively interact with the users to improve the search experience?
The existing conversational search systems mostly perform in reactive ways, which respond to the user for every query turn until the user terminates the search session. A more optimal interaction design is acting proactively, where the system can decide and engage in the conversation. This is also the vision of mixed-initiative interactions, while there is a lack of data to train and evaluate the systems. Another important element is to offer personalized search, which requires the conversation systems to consider appropriate aspects from the user profile for each query to help generate personalized content as the final response.

\noindent \textbf{Toward Practical Evaluation.} 
How to develop high-fidelity user simulations that simulate user behavior, including error recovery, adaptation, and evolving information goals, for practical system evaluation?
The existing datasets for evaluation are based on a strong assumption that all the previous turns are answered correctly when dealing with the current turn. This is far from the practical situation because the changed user behaviors will influence the later conversation flows. To address this open challenge, some efforts aim to develop user simulators to mimic real-world scenarios for practical evaluation. 
Besides, how to develop methodologies and comprehensive benchmarks to capture and evaluate long-term conversational success, task completion, information gain, and user satisfaction for practical conversational search scenarios is also valuable and promising.
Further exploration for benchmark developments and new evaluation paradigms is desirable for the research community.

\noindent \textbf{Informative Returns.} 
How to enrich the information within the returning response?
With the development of conversational search systems, the returning information might not be limited to clarification questions or generated responses. 
It might emphasize more on informative forms and proactive interaction.
The system could navigate the user toward different directions, e.g., calling external tools to tackle the request~\cite{schick2023tool}, planning a multi-step strategy as thought chains to decompose the request~\cite{wu2023autogen}, and collaborating with the environment to conduct agentic information retrieval~\cite{zhang2024agentic}. The key idea is to enlarge the available resources and capacity to help users reflect during the conversation and eventually achieve their information needs.

\noindent \textbf{New Paradigms of Conversational Search.}
How to integrate the developed capabilities of LLMs with the search engine toward a new paradigm of conversational search?
Given recent advances with LLMs and their adoption into day-to-day activities, the research community needs to reconsider what we are ultimately trying to achieve with conversational search. Understanding these goals would help contextualize how prior works attempted to solve the technical problems within their limitations, informing the algorithms, metrics, and exploration designs used. A new paradigm of conversational search by both inherits from the previous studies and considers the impacts of recent advances is desirable.




%







\balance
\bibliographystyle{IEEEtranN}
\bibliography{mybib_dblp}
%




%






\end{CJK}
\end{document}